%% file: main.tex
\documentclass[10pt,twocolumn,letterpaper]{article}

\usepackage{cvpr}              
\usepackage{graphicx}
\usepackage{amsmath}
\usepackage{amssymb}
\usepackage{booktabs}
\usepackage{graphicx}
\usepackage{algorithm}
\usepackage{algorithmic}
\usepackage{amsfonts}
\usepackage{multirow}
\usepackage{xcolor}
\usepackage{subcaption}
\usepackage{todonotes}
\usepackage{colortbl}
\usepackage[accsupp]{axessibility}

\newcommand{\cadapter}{\text{Conv-Adapter}\xspace}
\newcommand{\cellc}{\cellcolor{lightgray!50}}


\definecolor{cvprblue}{rgb}{0.21,0.49,0.74}
\usepackage[pagebackref,breaklinks,colorlinks,citecolor=cvprblue]{hyperref}

\def\paperID{7} 
\def\confName{CVPR}
\def\confYear{2024}

\title{Conv-Adapter: Exploring Parameter Efficient Transfer Learning for ConvNets}

\author{
Hao Chen$^1$\thanks{haoc3@andrew.cmu.edu} , Ran Tao $^1$, Han Zhang$^1$, Yidong Wang$^2$, Xiang Li$^1$, \\ 
Wei Ye$^2$, Jindong Wang$^3$, Guosheng Hu$^4$, Marios Savvides $^1$
\\
$^1$ Carneige Mellon University,  $^2$ Peking University, $^3$ Microsoft Research Asia, $^4$ Oosto
}

\begin{document}
\maketitle
\input{sec/0_abstract}    
\input{sec/1_intro}
{
    \small
    \bibliographystyle{ieeenat_fullname}
    \bibliography{main}
}

\input{sec/X_suppl}

\end{document}

%% file: sec/0_abstract.tex
\begin{abstract}
While parameter efficient tuning (PET) methods have shown great potential with transformer architecture on Natural Language Processing (NLP) tasks, their effectiveness with large-scale ConvNets is still under-studied on Computer Vision (CV) tasks. This paper proposes Conv-Adapter, a PET module designed for ConvNets. \cadapter is light-weight, domain-transferable, and architecture-agnostic with generalized performance on different tasks. 
When transferring on downstream tasks, Conv-Adapter learns tasks-specific feature modulation to the intermediate representations of backbones while keeping the pre-trained parameters frozen. By introducing only a tiny amount of learnable parameters, e.g., only $3.5\%$ full fine-tuning parameters of ResNet50.
It can also be applied for transformer-based backbones.
Conv-Adapter outperforms previous PET baseline methods and achieves comparable or surpasses the performance of full fine-tuning on $23$ classification tasks of various domains. It also presents superior performance on the few-shot classification with an average margin of $3.39$\%. Beyond classification, \cadapter can generalize to detection and segmentation tasks with more than $50\%$ reduction of parameters but comparable performance to the traditional full fine-tuning \footnote{Code is available at: \url{https://github.com/Hhhhhhao/Conv-Adapter/tree/main}}
\end{abstract}

%% file: sec/1_intro.tex
\section{Introduction}

As transfer learning \cite{thrun1998lifelong} thrives, large-scale foundation models gradually dominate deep learning over the last few years \cite{bommasani2021opportunities}. 
Fine-tuning has become the de-facto paradigm adapting a foundation model pre-trained on a pretext task to various downstream tasks for both Computer Vision (CV) and Natural Language Processing (NLP). 
Albeit its simplicity and prominence, fine-tuning has been posing challenges to development and deployment of the large-scale foundation models on downstream tasks with the drastic growth of computations and storage costs, as the parameter size increases from millions \cite{he2016resnet,howard2017mobilenets,tan2019efficientnet,ilija2020regnet} to billions \cite{devlin2018bert,brown2020language,fedus2021switch,dosovitskiy2020image,radford2021learningclip,liu2021Swin,liu2022convnet,liu2021swinv2}. 

Parameter efficient tuning (PET), as an alternative to traditional fine-tuning, has become prevalent in NLP \cite{pmlrv97houlsby19a,hu2021lora,li2021prefixtuning,lester2021powerprompt,he2022towards} for its efficiency and effectiveness.  
PET introduces a small number of learnable parameters to a pre-trained network, whose parameters are frozen, and learns the extra introduced parameters only. While attaining promising performance, especially for tasks of low-data regimes \cite{zhang2021differentiable,jia2022vpt,zhang2022NOAH}, PET modules for Convolutional Neural Networks (ConvNets),  the popular architectures for CV tasks, are still largely unstudied. 

\begin{figure}
    \centering
    \includegraphics[width=0.8\columnwidth]{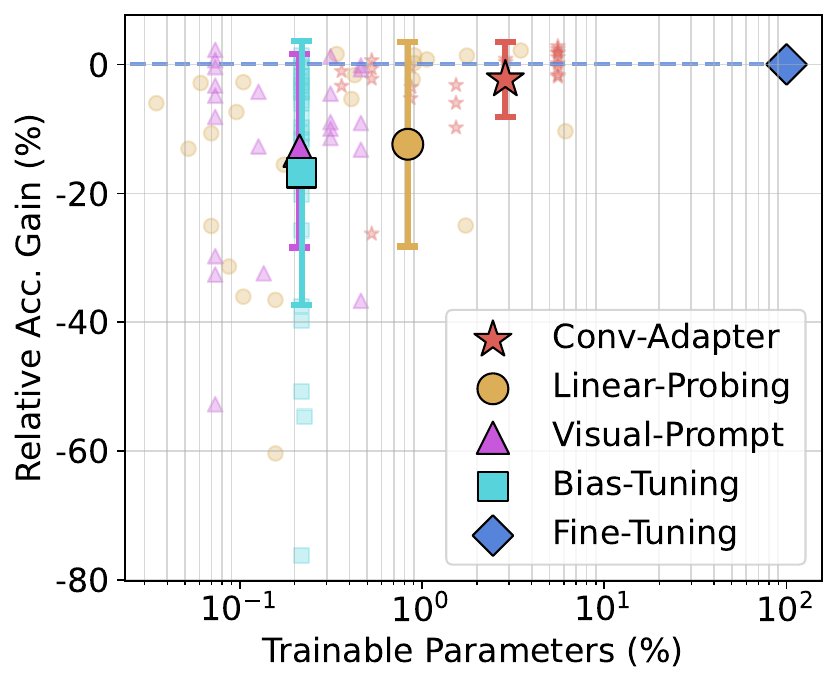}
    \caption{Performance of Conv-Adapter compared to other transfer learning methods on ResNet-50 BiT-M. We compute the relative performance gain w.r.t to fine-tuning and percentage of trainable parameters of the backbone (w/o linear head) on 23 image classification datasets from various domains to compute the results, with mean and standard deviation highlighted. Conv-Adapter achieves a superior trade-off between transfer accuracy and parameter efficiency.}
    \label{fig:r50_acc_param}
\end{figure}

Prior arts on fine-tuning ConvNets to multiple visual domains are restrictive in generalization and parameter efficiency.
Bias Tuning \cite{ben2022bias}, which tunes only the bias terms of the backbone, might fail on domains with significant distribution shifts from the pre-training tasks. 
Residual Adapter \cite{rebufficvpr2018resadapter} and TinyTL \cite{han2020nipstinyml} are mainly designed for small networks such as ResNet-26 \cite{he2016resnet} and MobileNet \cite{howard2017mobilenets,cai2018proxylessnas}. 
It is prohibitive to scale these previous designs to larger ConvNets \cite{liu2022convnet} or more diverse domains \cite{zhai2019largescale}. Besides, previous PET methods \cite{houlsby2019parameter,hu2021lora,li2021prefixtuning,lester2021powerprompt,he2022towards} are mainly designed with Transformer \cite{vaswani2017attention} architecture for NLP tasks \cite{devlin2018bert,brown2020language}.
However, it is not straightforward to apply Transformer-based PET to
ConvNets because Transformers tokenize and sequentialize the input and features, while ConvNets do not.
Recent works \cite{jia2022vpt,bahng2022visual,chen2022adaptformer} that attempt to use Prompt Tuning \cite{lester2021powerprompt} and Adapters \cite{houlsby2019parameter} on CV tasks are also designed for Vision Transformers rather than ConvNets. 
Furthermore, the downstream CV tasks are usually more diverse with a larger domain gap compared with NLP \cite{radford2021learningclip}.
These challenges motivate us to design the architecture and adapting scheme of PET for ConvNets, which could make it transferable to various CV tasks, including image classification, object detection, and semantic segmentation. 


In this work, we narrow the gap of PET between NLP and CV with the proposal of \textbf{Conv-Adapter} -- an adaption module that is light-weight, domain-transferable, and architecture-agnostic. 
Conv-Adapter learns task-specific knowledge on downstream tasks and adapts the intermediate features of each residual block in the pre-trained ConvNets.
It has a bottleneck structure consisting of depth-wise separable convolutions \cite{howard2017mobilenets} and non-linearity. 
Due to the variety of CV network architectures  
and tasks, we explore four adapting schemes of Conv-Adapter
combining two design perspectives - adapted representations and insertion form to verify the optimal tuning paradigm on ConvNets.
We find it is essential for Conv-Adapter to maintain the locality relationship when adapting intermediate feature maps for transferability. 
More importantly, Conv-Adapter can be formulated under the same mathematical framework as the PET modules used in the NLP field \cite{he2022towards}. 
Conv-Adapter outperforms previous PET baselines and achieves similar or even better performance to the traditional \emph{full} fine-tuning on $23$ cross-domain classification datasets with an average of $3.5$\% of the backbone parameters using ResNet-50 BiT-M \cite{alex2019big}, as shown in Fig. \ref{fig:r50_acc_param}. 
Conv-Adapter also well generalizes to object detection and semantic segmentation tasks with same-level performance to fully fine-tuning. 
To further understand Conv-Adapter, in addition, 
we empirically analyze the performance of Conv-Adapter with both the domain shifting of datasets and the network weights shifting brought by fine-tuning.
The core contributions of this work can be summarized as:
\begin{itemize}
    \item To our knowledge, we are the first to \emph{systematically} investigate the feasible solutions of general parameter-efficient tuning (PET) for ConvNets. This investigation can narrow the gap between NLP and CV for PET.  
    \item We propose Conv-Adapter, a light-weight and plug-and-play PET module, along with four adapting variants following two design dimensions - transferability and parameter efficiency. Meanwhile, we empirically justify several essential design choices to make Conv-Adapter effectively transferred to different CV tasks.
    \item Extensive experiments demonstrate the effectiveness and efficiency of Conv-Adapter. It achieves comparable or even better performance to full fine-tuning with only around 5\% backbone parameters. Conv-Adapter also well generalizes to detection and segmentation tasks that require dense predictions.
\end{itemize}

\section{Related Work}

\subsection{Parameter Efficient Tuning for Transformers}

Pre-trained Transformer models in NLP are usually of the size of billions of parameters \cite{devlin2018bert,brown2020language,fedus2021switch}, which makes fine-tuning inefficient as one needs to train and maintain a separate copy of the backbone parameters on each downstream task. 
Adapter \cite{houlsby2019parameter} is first proposal to conduct transfer with light-weight adapter modules.
It learns the task-specific knowledge and composes it into the pre-trained backbone \cite{pfeiffer2020adapterhub,pfeiffer2021adapterfusion} when adapting to a new task. 
Similarly, LoRA introduces trainable low-rank matrices to each layer of the backbone model to approximate parameter updates. Different from inserting adaption modules to intermediate layers, Prefix Tuning \cite{li2021prefixtuning} and Prompt Tuning \cite{lester2021powerprompt}, inspired by the success of textual prompts \cite{brown2020language,liu2021pretrain,radford2021learningclip}, prepend learnable prompt tokens to input and only train these tokens when transferring to a new task. 
More recently, a unified formulation of Adapter, LoRA, and Prefix Tuning is proposed in \cite{he2022towards}, where their core function is to adapt the intermediate representation of the pre-trained model by residual task-specific representation learned by tuning modules. 

Visual Prompt Tuning \cite{jia2022vpt} is a recent method adapting Prompt Tuning from NLP to Vision Transformers \cite{jia2022vpt}.
Bahng et. al. \cite{bahng2022visual} also explores visual prompts in input pixel space for adapting CLIP models \cite{radford2021learningclip} and makes connection with \cite{elsayed2018adversarial}.
While showing promising results on Transformers, visual prompts on ConvNets presents much worse transfer results \cite{jia2022vpt,bahng2022visual}, possibly due to the limited capacity of input space visual prompts. Conv-Adapter can adapt the intermediate features thus has larger capacity.

\subsection{Transfer Learning for ConvNets}

While there is no straightforward approach to applying previous PET methods designed for Transformers directly on ConvNets, several attempts have been made in prior research. 
BatchNorm Tuning \cite{mudrakarta2018k} and Bias Tuning \cite{ben2022bias} only tune the batchnorm related terms or the bias terms of the pre-trained backbone. 
Piggyback \cite{mallya2018piggyback} instead learns weight masks for downstream tasks while keeping the pre-trained backbone unchanged. 
They all have limited transferability and update partial parameters of the backbone.

More related to our work, Residual Adapter \cite{rebufficvpr2018resadapter} explores inserting an extra convolutional layer of kernel size 1 to each convolutional layer in pre-trained ResNet-26 \cite{he2016resnet}, either in parallel or in sequential, to conduct the multi-domain transfer. 
Similarly, TinyTL introduces extra residual blocks to MobileNet \cite{howard2017mobilenets,cai2018proxylessnas} for memory efficient on-device learning. 
Guo et. al. \cite{guo2019depthwise} proposes re-composing a ResNet with depth-wise and point-wise convolutions, and re-training only the depth-wise part during fine-tuning. 
RepNet \cite{yang2022repnet} exploits a dedicated designed side network to re-program the intermediate features of pre-trained ConvNets. 
Conv-Adapter differs from previous methods with a design that considers parameter efficiency and transferability from the internal architectures and adapting schemes. Besides, the proposed Conv-Adapter does not require tuning any backbone parameters to achieve comparable performance to fine-tuning.

\section{Method}

\subsection{Preliminaries}

Parameter efficient tuning (PET) methods \cite{houlsby2019parameter,hu2021lora,li2021prefixtuning,lester2021powerprompt,jia2022vpt} introduce learnable adapting modules plugged into the backbone that is frozen during tuning.
From a unified point of view, the core function of the adaption modules is to learn task-specific feature modulations on originally hidden representations in the pre-trained backbone \cite{he2022towards}.
Specifically, considering an intermediate hidden representation $\mathbf{h}$ generated by a layer or a series of layers with input $\mathbf{x}$ in a pre-trained network, the PET adaption module learns $\Delta \mathbf{h}$ and updates $\mathbf{h}$ as: 
\begin{equation}
    \mathbf{h} \xleftarrow{} \mathbf{h} + \mathbf{\alpha} \cdot  \Delta \mathbf{h},
\label{eq:formulation}
\end{equation}
where $\mathbf{\alpha}$ could be a scalar \cite{hu2021lora} or a gating function \cite{li2021prefixtuning}. Previous PET methods in NLP mainly follow a similar functional form for constructing $\Delta \mathbf{h}$ -- down-sampling projection,  non-linearity, and up-sampling projection. 
However, they differ in 1) implementation (architecture) - the form of the projections and non-linearity, and 2) the adapting scheme - which $\mathbf{h}$ in the model to adapt and compute $\Delta \mathbf{h}$ from which representation.
These differences characterize the adaptation to new tasks and robustness to out-of-distribution evaluation \cite{li2021prefixtuning}. 

It is non-trivial to design effective PET methods for ConvNets because previous PET modules are mainly developed on Transformers rather than ConvNets.
Besides, the components of the architecture and computation dynamics of ConvNets and Transformers are inherently different.
Following the unified formulation of PET methods in Eq. (\ref{eq:formulation}),
we propose \textbf{\cadapter}.
We construct the $\Delta \mathbf{h}$ of \cadapter similarly to previous PET methods and design the adaption architecture and scheme on ConvNets from the perspective of transferability and parameter efficiency.

\subsection{Motivation}



Before delving into the details of our design, we identify the essential difficulty that prevents utilizing prior arts directly on ConvNets as an adaption module and thus inspires us to propose \cadapter. Conventionally, for ConvNets, $\mathbf{h}$ and $\Delta \mathbf{h}$ are usually 3-dimensional structural features maps belonging to $\mathbb{R}^{C \times H \times W}$ with $C$ being the channel dimension and $H \times W$ being the spatial size of the feature maps. 

The difference in intermediate feature and processing dynamics poses obstacles to transferability. For Transformers, $\mathbf{h}$ is whereas 2-dimensional sequential features in $\mathbb{R}^{L \times D}$ where $L$ is the sequence length and $D$ is the feature dimension. 
Previous PET modules for Transformers compute $\Delta \mathbf{h}$ in various forms, e.g., linear layers over $\mathbf{h}$ \cite{houlsby2019parameter} and self-attention over additional input prompts \cite{li2021prefixtuning,lester2021powerprompt,jia2022vpt}.
They can all process the sequential features globally with long-range dependencies as the computing blocks in Transformers. 
Although it is possible to apply linear layers, or equivalently $1 \times 1$ convolutional layers \cite{rebufficvpr2018resadapter}, to adapt the feature maps of ConvNets, it is yet intuitive that this might produce inferior transfer performance due to the \emph{loss of locality}, which is encoded in the structural features maps by convolutions of kernel size larger than 1. The \emph{loss of locality} results in a radical mismatch of the receptive field in $\Delta \mathbf{h}$ and $\mathbf{h}$, which might be destructive when adapting ConvNets on tasks with significant domain shifts.
Apart from the receptive field mismatch, the spatial size of feature maps in ConvNets also significantly affects the transferability of adaption.
Earlier attempts to use adapters to transfer ConvNets usually downsample the feature's spatial size for memory and parameter efficiency. 
However, for CV tasks beyond image classification like segmentation, the spatial size matters for achieving good results \cite{shelhamer2017fully,chen2017rethinking}.

In summary, it is crucial to design the architecture and adapting scheme of the PET module computing $\Delta \mathbf{h}$ for ConvNets to have the same spatial size of feature maps and the same receptive field of convolutions for transferability. 


\begin{figure}[!t]
    \centering
    \includegraphics[width=0.65\columnwidth]{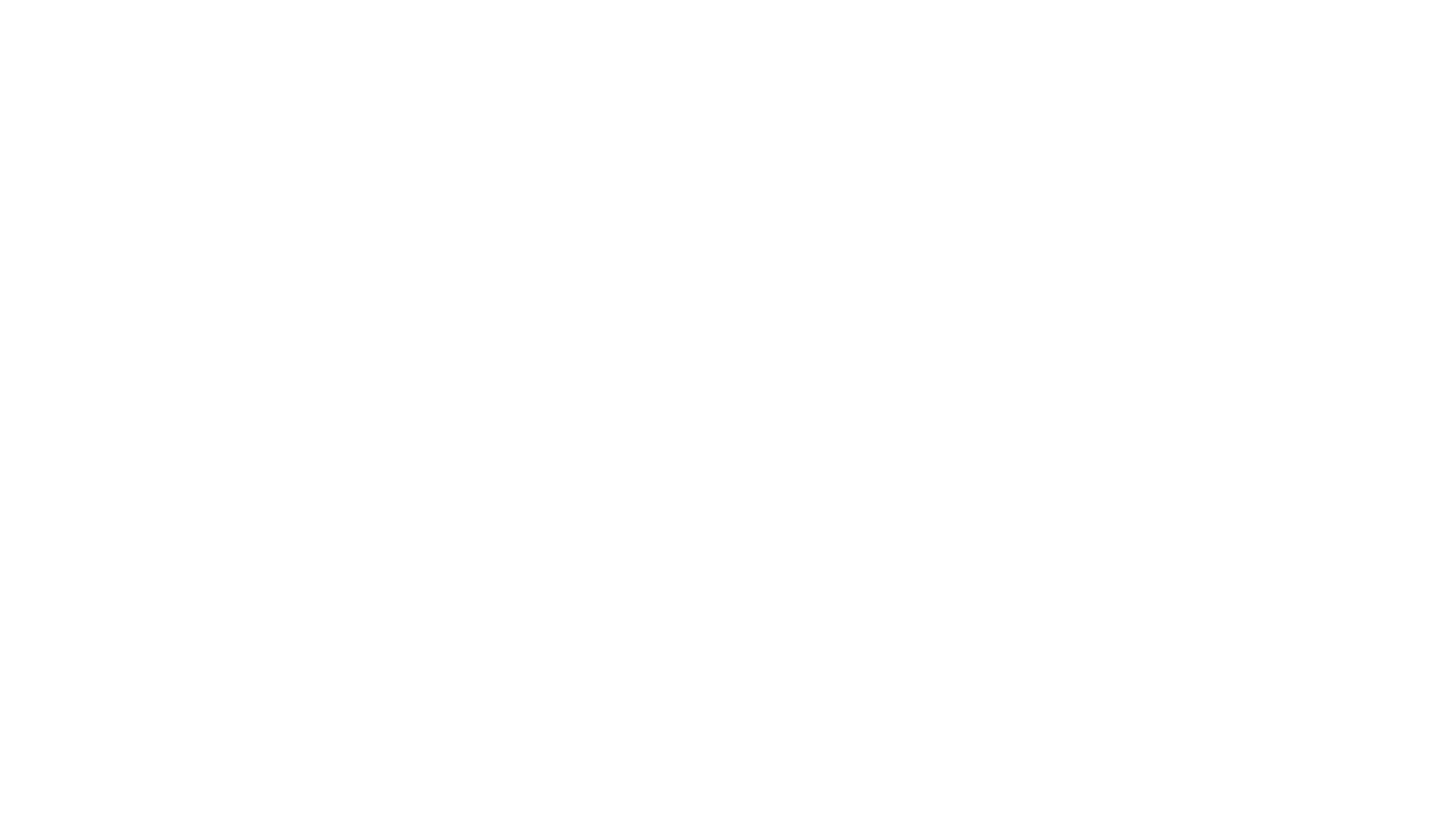}
    \caption{Architecture of Conv-Adapter, which has a bottleneck composed of depth-wise separable convolutions with non-linearity activation. $C_{in}$, $C_{out}$, $H$, $W$ is set to keep the same as in backbone. $\boldsymbol{\alpha}$ and $\gamma$ are hyper-parameters to tune.}
    \label{fig:adapter_arch}
\end{figure}

\begin{figure*}[!t]
    \centering
    \includegraphics[width=0.95\textwidth]{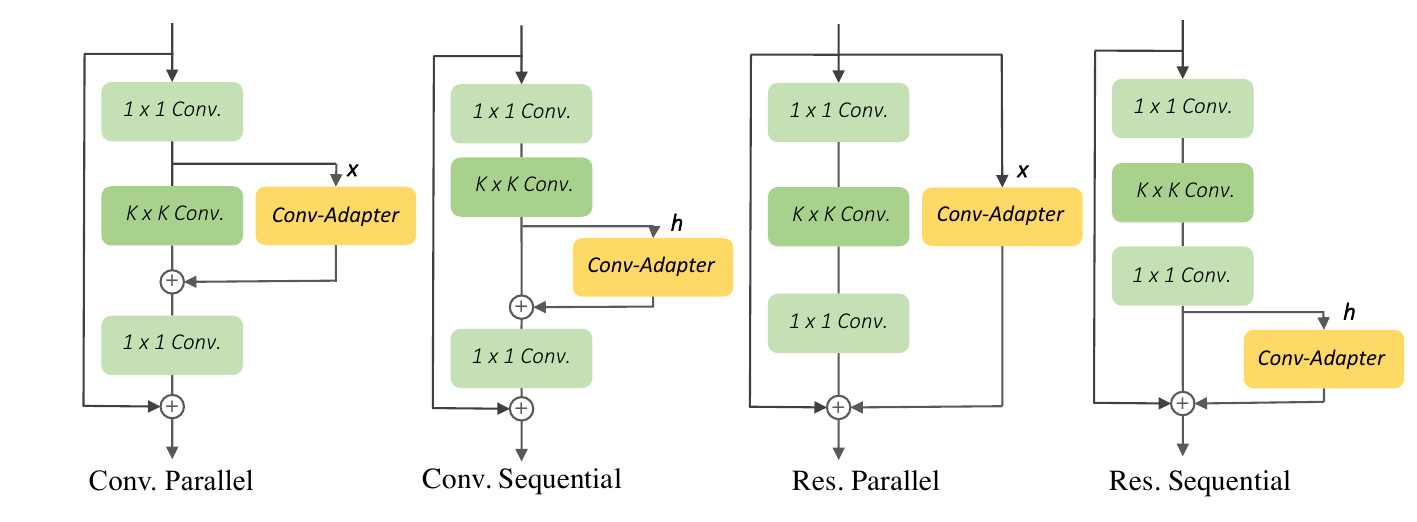}
    \caption{Four adapting schemes of Conv-Adapter to ResNet50: Convolution Parallel, Convolutional Sequential, Residual Parallel, and Residual Sequential. The schemes differ regarding the position of of the modified representation and corresponding insertion form. Other networks can be adapted similarly following the illustration. Green modules are frozen during fine-tuning.}
    \label{fig:adapt_scheme}
\end{figure*}

\subsection{Architecture of Conv-Adapter}

Given the above challenges, we design our \textbf{\cadapter} as a bottleneck structure, which is also widely used by PET methods of NLP tasks \cite{he2016resnet,houlsby2019parameter}. However, our \textbf{\cadapter} designs the bottleneck, particularly for ConvNets. Precisely, it consists of two convolutional layers with a non-linearity function in-between. 
The first convolution conducts channel dimension down-sampling with a kernel size similar to that of the adapted blocks, whereas the second convolution projects the channel dimension back. 
For simplicity, we adopt the same activation function used in the backbone as the non-linearity at the middle of the bottleneck. 
The effective receptive field of the modulated feature maps produced by Conv-Adapter is thus similar to that of the adapted blocks in the backbone. 
We do not change the spatial size of the feature maps for better transferability on dense prediction tasks.
We adopt the depth-wise separable convolutions \cite{howard2017mobilenets} for Conv-Adapter to reduce the parameter size further. 

\figurename~\ref{fig:adapter_arch} illustrates  our \cadapter architecture. 
Formally, let the input feature map to the adapted blocks of the ConvNets be $\mathbf{z} \in \mathbb{R}^{C_{in} \times H \times W}$ and the output feature maps be $\mathbf{h} \in \mathbb{R}^{C_{out} \times H \times W}$, where $C_{in}$ and $C_{out}$ are the channel dimension of the input and output to the adapted blocks respectively.
Assuming the spatial size $H \times W$ of the feature maps does not change along these blocks, we set the learnable weight as $\mathbf{W}_{down} \in \mathbb{R}^{\frac{C_{in}}{\gamma} \times \gamma \times K \times K}$ for the depth-wise convolution and $\mathbf{W}_{up} \in \mathbb{R}^{C_{out} \times \frac{C_{in}}{\gamma} \times 1 \times 1}$ for the point-wise convolution in Conv-Adapter, with the non-linearity denoted as $f$. We use a compression factor of $\gamma$ to denote the down-sampling in the channel dimension, where $\gamma$ is a hyper-parameter tuned for each task.
Mathematically, Conv-Adapter computes $\Delta \mathbf{h}  \in \mathbb{R}^{C_{out} \times H \times W}$ as:
\begin{equation}
    \Delta \mathbf{h} = \left( \mathbf{W}_{up} \otimes f(\mathbf{W}_{down} \hat{\otimes}  \mathbf{z}) \right),
\end{equation}
where $\otimes$ and $\hat{\otimes}$ denotes point-wise  and depth-wise convolution, respectively. 
To allow the modulation $\Delta \mathbf{h}$ to be more flexibly composed into $\mathbf{h}$, we set $\boldsymbol{\alpha}$ in Eq. (\ref{eq:formulation}) as a learnable scaling vector in $\mathbb{R}^{C_{out}}$, which is initialized as ones. 
The ablation study on design choices is presented in Sec.~\ref{sec:ablation}.

\subsection{Adapting ConvNets with Conv-Adapter}

After setting the architecture of Conv-Adapter, 
we discuss the scheme to adapt a variety of ConvNets.
Previous PET methods insert the adapting modules to Self-Attention blocks, Feed-Forward blocks, or both \cite{he2022towards} of Transformers, which have a relatively unified architecture.
In contrast, modern ConvNets usually stacks either residual blocks \cite{szegedy2015going,he2016resnet,zhang2020resnest} or inverted residual blocks \cite{howard2017mobilenets,tan2019efficientnet,tan2021efficientnetv2,liu2022convnet}, which consists of a series of convolutional layers (and sometimes pooling layers) and a residual identity branch, making it more difficult to use a single adapting scheme to various architectures. 

To explore the effective adapting schemes of using Conv-Adapter to tune a ConvNet, we study it mainly from two perspectives, similar to 
\cite{he2022towards}, 1) the location of adaptation in pre-trained ConvNets -- which intermediate representation $\mathbf{h}$ to adapt, and 2) the insertion form of Conv-Adapter -- how to set the input $\mathbf{z}$ to Conv-Adapter to compute $\Delta \mathbf{h}$. 
From the location perspective, we study plugging Conv-Adapter to each (inverted) residual block \cite{han2020nipstinyml} or to each functioning $K \times K$ convolutional layer within a residual block \cite{guo2019depthwise}. 
From the insertion perspective, Conv-Adapter can be inserted either in parallel or in sequential to the modified components, with the input to Conv-Adapter being $\mathbf{x}$, the input to the modified components, or being $\mathbf{h}$ itself, respectively. 
Combining the design dimension from these two perspectives, we propose 4 variants of adapting schemes with Conv-Adapter: \textbf{Convolution Parallel}, \textbf{Convolution Sequential}, \textbf{Residual Parallel}, and \textbf{Residual Sequential}. 

Taking the bottleneck residual block of ResNet-50 \cite{he2016resnet} as an example, we demonstrate the proposed designs  
in Fig. \ref{fig:adapt_scheme}. 
As $1 \times 1$ convolution layer can only transfer channel-wise information, we thus design the adapting of functional convolutions, i.e., intermediate $K \times K$ convolutions, to keep locality sensitive. 
On the contrary, adapting the whole residual block considers the transferring of pre-trained knowledge carried by $1 \times 1$ convolutions.
Intuitively, adapting the whole residual blocks has a larger capacity for modulating task-specific features than adapting only $K \times K$ convolution but may introduce more parameters.
Plugging \cadapter stage-wisely is not considered as it is impractical to make the receptive field of \cadapter similar to the adapted stage with only two convolutions. 
It needs a more sophisticated design on not only the Conv-Adapter architecture but also the adaptation location \cite{yang2022repnet}, and we empirically find that stage-wise adaptation produces inferior performance and requires much more parameters.
Conv-Adapter is flexible to be inserted into every residual block of the ConvNet backbone for transferability of features from different depths, as in \cite{rebufficvpr2018resadapter,mallya2018piggyback}.
Other backbones such as ConvNext \cite{liu2022convnet}, and even Swin-Transformer \cite{liu2021Swin} can be adapted following the same guideline (see experiments).

\section{Experiments}
\label{sec:exp}

This section verifies the transferability and parameter efficiency of \cadapter from various aspects, including image classification, few-shot classification, object detection, and semantic segmentation. Additionally, we provide an ablation study of \cadapter for its design choices and an analysis of its performance.

\subsection{Transferability of Conv-Adapter} 
\label{sec:trans_cls}

\subsubsection{Setup}
We first evaluate the transferability of Conv-Adapter on classification tasks. We experiment on two benchmarks: VTAB-1k \cite{zhai2019largescale} and FGVC. 
VTAB-1k includes 19 diverse visual classification tasks, which are grouped into three categories: \textit{Natural}, \textit{Specialized}, and \textit{Structured} based on the domain of the images. Each task in VTAB-1k contains 1,000 training examples. FGVC consists of 4 \textit{Fine-Grained Visual Classification} tasks: CUB-200-2011 \cite{WahCUB2002011}, Stanford Dogs \cite{KhoslaYaoJayadevaprakashFeiFeiFGVC2011}, Stanford Cars \cite{KrauseStarkDengFeiFei3DRR2013}, and NABirds \cite{van2015building}. 

For evaluation, we compare the 4 variants of Conv-Adapter with full fine-tuning (FT) and 3 baseline methods: linear probing (LP), bias tuning (Bias) \cite{han2020nipstinyml}, and visual prompt tuning (VPT) \cite{jia2022vpt, bahng2022visual}. We test each method on ResNet50 \cite{he2016resnet,alex2019big} with ImageNet21k pre-training. To find the optimal hyper-parameters of Conv-Adapter (and baseline methods), we conduct a grid search of the learning rate, weight decay, and compression factor $\gamma$ for each dataset using the validation data split from training data for both benchmarks. 
For VTAB-1k, we use the recommended optimal data augmentations in \cite{zhai2019largescale}, rather than solely Resize and Centre Crop as in \cite{bahng2022visual,zhang2022NOAH}. We find the recommended augmentations produces better results for full-tuning. For FGVC, we use RandomResized Crop with a minimum scale of 0.2 and Horizontal Flip \cite{2015inception} as augmentation. 
Mores details of the hyper-parameters are shown in Appendix. 

\begin{table}[t!]
\centering
\caption{Performance of Conv-Adapter adapting schemes on ResNet-50 BiT-M. Each setting includes three runs and averaged top-1 accuracy ($\%$) over datasets and the averaged total trainable parameters (M) over all datasets are reported.
We compare proposed variants of Conv-Adapter (in gray) with full Fine-Tuning (FT), Linear Probing (LP), Bias Tuning (Bias), and Visual Prompt Tuning (VPT). 
We report the number of wins of $(\cdot)$ for each method compared to FT.
\textbf{Bold} and \underline{underline} refer to the top and second result separately.
}
\label{tab:transferability}
\resizebox{0.9\columnwidth}{!}{%
\begin{tabular}{@{}l|r|ccccc@{}}
\toprule[1pt]
  \multicolumn{1}{c|}{\multirow{2}{*}{Tuning}} &
  \multicolumn{1}{c|}{\multirow{2}{*}{\# Param.}} &
  \multirow{2}{*}{FGVC} &
  \multicolumn{3}{c}{VTAB-1k} \\ \cmidrule(l){4-6} 
 &
  \multicolumn{1}{c|}{} &&
  Natural &
  Specialized &
  Structured \\ \midrule
\# Tasks
& -
& 4 & 7 & 4 & 8 
\\ \midrule

                                FT                & 23.89 & 83.46          & 72.19 & \textbf{85.86} & \textbf{66.72}  \\ \cmidrule(l){1-6} 
                                LP                & 0.37  & 75.44 (1)             & 67.42 (4) & 81.42 (0) & 37.92 (0)  \\
                                Bias              & 0.41  & 64.98 (0)           & 66.06 (4) & 80.34 (0) & 32.18  (0) \\
                                VPT               & 0.42  & 74.79 (1)             & 65.43 (2) & 80.35 (0) & 37.64 (0) \\  \cmidrule(l){1-6} 
                                \cellc Conv. Par.        & \cellc 0.85  &\cellc \underline{83.77 (3)}     & \cellc \textbf{72.60 (5)} & \cellc 84.21 (1) & \cellc 56.70 (1) &  \\
                                \cellc Conv. Seq.        & \cellc 0.87  &\cellc 79.68 (2)             & \cellc \underline{72.28 (4)} &\cellc 83.85 (0) & \cellc 58.50 (1) &  \\
                                \cellc Res. Par.         & \cellc 8.21  & \cellc \textbf{84.24 (3)}             & \cellc 71.75 (4) & \cellc 84.70 (0) & \cellc \underline{61.34 (1)} &  \\
                                \cellc Res. Seq.         & \cellc 3.53  & \cellc 83.45 (2)             & \cellc 71.74 (4) & \cellc \underline{84.84 (0)} & \cellc 61.33 (2) &  \\ 
\bottomrule[1pt]
\end{tabular}%
}
\end{table}

\subsubsection{Results and Discussion}
Results are reported in Tab.~\ref{tab:transferability}.
Conv-Adapter not only demonstrates significant improvements over the baseline methods, but also achieves the same level of performance or even surpasses their fine-tuning counterparts on all domains evaluated, by introducing only around \textbf{3.5\%} of full fine-tuning parameters for ResNet-50.
Notably, there is a considerable performance gap, i.e., an improvement of \textbf{23.44\%}, of Conv-Adapter over previous baseline methods on \textit{Structured} datasets of VTAB-1k.

One can observe that the proposed four variants of Conv-Adapter all achieve comparable performance compared to full fine-tuning.
Among the four variants, \textbf{Convolution Parallel} achieves the best trade-off between performance and parameter efficiency.
On the evaluated classification tasks, inserting Conv-Adapter in parallel generally outperforms inserting sequentially. 
In terms of the modified representation, one can find that, on most of the datasets, adapting only the $K \times K$ convolutions of ResNet-50 can achieve performance close to fine-tuning. 
However, on \textit{Structured} datasets, adapting whole residual blocks is far better than adjusting only the middle convolutions with more parameters, 
demonstrating the superior capacity of adjusting residual blocks when there is a more significant domain gap.

\subsection{Universality of Conv-Adapter}\label{sec:uni_cls}

\subsubsection{Setup} 
We evaluate the universality of \cadapter on classification tasks in this section, where \cadapter is inserted to various ConvNets architectures with different pre-training. 
We adopt the simple yet effective adapting scheme -- Convolution Parallel, and mainly compare it with full fine-tuning. 
More specifically, we adopt ImageNet-21k pre-trained ResNet50 \cite{alex2019big}, ConvNext-B and ConvNext-L \cite{liu2022convnet}, and even Swin-B and Swin-L \cite{liu2021Swin}. 
Apart from ImageNet-21k, we evaluate ImageNet-1K, CLIP \cite{he2020moco}, and MoCov3 \cite{chen2021mocov3} pre-training.
Similarly, we conduct a hyper-parameter search on the validation set, and report the accuracy on the test set of FGVC and VTAB-1k.
Model details are shown in Appendix.

\subsubsection{Results and Discussion} 
We present the results in Tab. \ref{tab:universality}. On various ImageNet-21k pre-trained ConvNets, Conv-Adapter demonstrates its universality with comparable performance to fine-tuning. 
For large models such as ConvNext-L and Swin-L, conducting traditional fine-tuning requires training nearly 196M parameters, whereas Conv-Adapter improves the parameter efficiency with only 7.8\% and 4.5\% of the fine-tuning parameters on ConvNext-L and Swin-L respectively. 
Although the transfer performance of Conv-Adapter on ImageNet-1k pre-trained models is more limited, compared to ImageNet-21k pre-training, Conv-Adapter still demonstrates its superior parameter efficiency and shows improvement over fine-tuning on several tasks.
For the CLIP vision models, Conv-Adapter consistently outperforms fine-tuning on Structured tasks of VTAB-1k. 
We observe a performance gap of Conv-Adapter on MoCov3 pre-trained \cite{chen2021mocov3}, and we argue this is possibly due to the difference in feature space of self-supervised and supervised models in CV \cite{jia2022vpt}.

\begin{table}[t!]
\centering
\caption{Comparing Conv-Adapter (CA) with full Fine-Tuning (FT) using various backbone architectures of different pre-training. Each setting includes three runs and averaged top-1 accuracy ($\%$) over datasets and the averaged total trainable parameters (M) over all datasets are reported. We report the number of wins of $(\cdot)$ for each method in compared to FT. \textbf{Bold} indicates the best results.}
\label{tab:universality}
\resizebox{\columnwidth}{!}{%
\begin{tabular}{@{}c|c|c|r|cccc@{}}
\toprule[1pt]
\multirow{2}{*}{Pre-train} &
  \multicolumn{1}{c|}{\multirow{2}{*}{Backbone}} &
  \multicolumn{1}{c|}{\multirow{2}{*}{Tuning}} &
  \multicolumn{1}{c|}{\multirow{2}{*}{\# Param.}} &
  \multicolumn{1}{c}{\multirow{2}{*}{FGVC}} &
  \multicolumn{3}{c}{VTAB-1k} \\ \cmidrule(l){6-8} 
 &
  \multicolumn{1}{c|}{} &
  \multicolumn{1}{c|}{} &
  \multicolumn{1}{c|}{} &
  \multicolumn{1}{c}{} &
  \multicolumn{1}{c}{Natural} &
  \multicolumn{1}{c}{Specialized} &
  \multicolumn{1}{c}{Structured} \\ \midrule
\# Tasks & & & & 4 & 7 & 4 & 8 \\ \midrule
\multirow{10}{*}{\begin{tabular}{@{}c@{}}ImageNet \\ 21k\end{tabular} } & \multirow{2}{*}{\begin{tabular}{@{}c@{}}ResNet50 \\ BiT-M\end{tabular}}   & FT           & 23.89 & 83.46 & 72.19 & \textbf{85.86} & \textbf{66.72} \\
                             &                             & \cellc CA           & \cellc 0.85  & \cellc \textbf{83.77 (3)} & \cellc \textbf{72.60 (5)}  & \cellc 84.21 (1) & \cellc 56.70 (1)  \\ \cmidrule(l){2-8} 
                             & \multirow{2}{*}{ConvNext-B} & FT           & 87.75 & \textbf{89.48} & \textbf{81.59} & \textbf{87.32} & \textbf{65.77} \\
                             &                             & \cellc CA           & \cellc 6.83  & \cellc 89.28 (1) & \cellc 80.62 (4) & \cellc 86.29 (0) & \cellc 64.88 (2) \\ \cmidrule(l){2-8} 
                             & \multirow{2}{*}{ConvNext-L} & FT       & 196.50    & 90.64  & \textbf{82.25} & \textbf{87.94} & \textbf{67.65}  \\
                             &                             & \cellc CA           & \cellc 15.52  & \cellc \textbf{90.69 (3)} & \cellc 81.7 (2) & \cellc 86.85 (0) & \cellc 64.98 (3) \\ \cmidrule(l){2-8} 
                             & \multirow{2}{*}{Swin-B}     & FT           & 86.92 & \textbf{90.01} & 78.65 & \textbf{87.59} & \textbf{64.69} \\
                             &                             & \cellc CA           & \cellc 4.98  & \cellc 88.55 (1) & \cellc \textbf{80.00 (4)}    & \cellc 85.84 (0) & \cellc 62.57 (2) \\ \cmidrule(l){2-8}
                             & \multirow{2}{*}{Swin-L}     & FT           & 195.27 & \textbf{91.04} & 80.64 & \textbf{87.85} & \textbf{66}  \\
                             &                             & \cellc CA           & \cellc 8.86 & \cellc 90.54 (2) & \cellc \textbf{81.39 (3)} & \cellc 86.29 (1) & \cellc 63.19 (2) \\ 
                             \midrule
\multirow{6}{*}{\begin{tabular}{@{}c@{}}ImageNet \\ 1k\end{tabular} }  & \multirow{2}{*}{ResNet50}   & FT           & 23.87 & \textbf{85.84} & \textbf{67.15} & \textbf{83.53} & \textbf{53.32} \\
                             &                             & \cellc CA           & \cellc 0.72  & \cellc 83.48 (0) & \cellc 64.20 (0)  &\cellc  81.33 (1) & \cellc 52.74 (2) \\ \cmidrule(l){2-8} 
                             & \multirow{2}{*}{ConvNext-B} & FT           & 87.75 & \textbf{88.95} & 74.51 & \textbf{85.33} & 61.34 \\
                             &                             & \cellc CA           & \cellc 10.82 & \cellc 87.84 (1) & \cellc \textbf{74.72 (4)} & \cellc 84.29 (0) & \cellc \textbf{63.77 (2)} \\ 
                             \midrule
\multirow{4}{*}{CLIP}        & \multirow{2}{*}{ResNet50}   & FT           & 38.50 & \textbf{81.38} & \textbf{58.53} & \textbf{80.8}  & 57.18 \\
                             &                             & \cellc CA           & \cellc 2.23  & \cellc 76.64 (0) &\cellc 56.33 (3) & \cellc 79.12 (0) & \cellc \textbf{58.96 (4)} \\ \cmidrule(l){2-8} 
                             & \multirow{2}{*}{ResNet50x4} & FT           & 87.17 & \textbf{84.23} & \textbf{65.71} & \textbf{82.22} & 58.84 \\
                             &                             & \cellc CA           & \cellc 6.14  & \cellc 82.71 (0) & \cellc 62.54 (2) & \cellc 80.72 (1) & \cellc \textbf{59.10 (4)}  \\  \midrule
\multirow{2}{*}{MoCov3}       & \multirow{2}{*}{ResNet50}   & FT           & 23.87  & \textbf{83.92} & \textbf{66.25} & \textbf{83.89}  & \textbf{60.26} \\
                             &                             & \cellc CA           & \cellc 0.89 &\cellc 79.69 (0)  &\cellc 65.31 (3)  &\cellc 81.59 (0) & \cellc 53.87 (1) \\ 
\bottomrule
\end{tabular}%
}
\end{table}

\subsection{Few-Shot Classification}

\subsubsection{Setup} 
PET methods usually present superior performance for tasks with low-data regimes \cite{li2021prefixtuning,he2022towards}.
We thus evaluate Conv-Adapter on few-shot classification using ImageNet-21k pre-trained ResNet50 Bit-M \cite{alex2019big} and ConvNext-B \cite{liu2022convnet}. 
We evaluate 5 FGVC datasets using 1, 2, 4, 8 shots for each class following following previous studies \cite{radford2021learningclip,jia2022vpt,zhang2022NOAH} including Food101 \cite{bossard14food101}, Oxford Flowers \cite{nilsback2006visual}, Oxford Pets \cite{parkhi12a}, Stanford Cars \cite{KrauseStarkDengFeiFei3DRR2013}, and Aircraft \cite{maji13finegrained}. Averaged top-1 accuracy is reported in Tab.~\ref{tab:few_shot}.
We search from the same range as before and adopt the same augmentations as for FGVC tasks.
The detailed hyper-parameters and more results for each dataset are in Appendix. 

\subsubsection{Results and Discussion} 
Compared with Fine-tuning, Conv-Adapter boosts few-shot classifications with an average 3.39\% margin over different shots using only around 5\% trainable parameters. Especially for 1/2-shot cases, Conv-Adapter shows supreme performance compared with Fine-tuning and VPT \cite{jia2022vpt} (11.07\% on 1-shot and 6.99\% on 2-shot with larger architecture ConvNext-B). 
Meanwhile, Conv-Adapter provides a better accuracy-efficiency trade-off than Visual Prompt Tuning on few-shot classifications. It surpasses VPT with an average margin of 1.35\% with ResNet50 Bit-M and 3.69\% with ConvNext-B. In the 8-shot case, VPT drops around 8\% performance compared with Fine-tuning due to limited capacity, while Conv-Adapter can achieve comparable or better performance to Fine-tuning and maintain parameter efficiency.


\begin{table}[t!]
\centering
\caption{Few-shot classification: the average Top-1 accuracy over 5 FGVC datasets, with 1, 2, 4, 8 shots. We compare Conv-Adapter (CA), Visual Prompt Tuning (VPT), and full Fine-Tuning (FT). \textbf{Bold} indicates the best results. 
}
\label{tab:few_shot}
\resizebox{0.75\columnwidth}{!}{%
\begin{tabular}{@{}c|l|l|rrrr@{}}
\toprule[1pt]
\multicolumn{1}{c|}{Backbone} &
  \multicolumn{1}{c|}{Tuning} &
  \multicolumn{1}{c|}{\# Param} &
  \multicolumn{1}{c}{1} &
  \multicolumn{1}{c}{2} &
  \multicolumn{1}{c}{4} &
  \multicolumn{1}{c}{8} \\ \midrule
\multirow{3}{*}{\begin{tabular}[c]{@{}c@{}}ResNet50\\ Bit-M\end{tabular}} &
  FT &
  23.72 &
  29.30 &
  38.96 &
  50.09 &
  61.27 \\
                            & VPT   & 0.24  & 32.56          & 42.18          & 52.21          & 59.37          \\
                            & \cellc CA   & \cellc 1.02  & \cellc \textbf{34.31} & \cellc \textbf{43.55} & \cellc \textbf{52.43} & \cellc \textbf{61.42} \\ \midrule
\multirow{3}{*}{ConvNext-B} & FT & 87.68 & 36.34          & 48.83          & \textbf{63.69}          & \textbf{76.91}          \\
                            & VPT   & 0.13  & 42.25          & 51.85          & 62.89          & 69.04          \\
                            & \cellc CA   & \cellc 4.6   & \cellc \textbf{47.41} & \cellc \textbf{55.82} & \cellc 63.25          & \cellc 74.29          \\ \bottomrule
\end{tabular}%
}
\end{table}

\subsection{Object Detection and Semantic Segmentation}

\begin{table}[t!]
    \centering
    \caption{Object detection \& Semantic Segmentation results. We report the results of fine-tuning and \cadapter with the Residual Parallel scheme.}
    \resizebox{0.75\columnwidth}{!}{%
    \begin{tabular}{@{}c|c|c|ccc@{}}
    \toprule[1pt]
      \multicolumn{6}{c}{Object Detection with Faster-RCNN} \\
      \multicolumn{1}{c|}{Backbone} &
      \multicolumn{1}{c|}{Tuning} & 
      \multicolumn{1}{c|}{\# Param} &
      \multicolumn{1}{c}{AP} &
      \multicolumn{1}{c}{AP\textsubscript{50}} &
      \multicolumn{1}{c}{AP\textsubscript{75}} \\ \midrule
      \multirow{2}{*}{ResNet50}     & FT  & 41.53 & 38.1 & 59.7 & 41.5 \\
       &\cellc CA & \cellc  35.72 & \cellc  \textbf{38.4} & \cellc  \textbf{61.1} & \cellc  \textbf{41.5}  \\ \midrule
     \multirow{2}{*}{ConvNeXt-B}  &    FT & 67.09 & \textbf{45.2} & \textbf{67.2} & \textbf{49.9}    \\
     & \cellc  CA & \cellc  24.62 & \cellc  41.9 & \cellc  64.5 & \cellc  45.7 \\
      \bottomrule
    \multicolumn{6}{c}{} \\
    
    \toprule[1pt]
      \multicolumn{6}{c}{Semantic Segmentation with UPerNet} \\
      \multicolumn{1}{c|}{Backbone} &
      \multicolumn{1}{c|}{Tuning} & 
      \multicolumn{1}{c|}{\# Param (M)} &
      \multicolumn{3}{c}{mIoU} \\ \midrule
      \multirow{2}{*}{ResNet50}     & FT  & 66.49 &  \multicolumn{3}{c}{42.1}  \\
       & \cellc  CA & \cellc  45.65 &  \multicolumn{3}{c}{\textbf{\cellc 43.0}} \\ \midrule
     \multirow{2}{*}{ConvNeXt-B} &  FT & 81.87 &  \multicolumn{3}{c}{\textbf{48.7}}   \\
     & \cellc  CA &\cellc   39.40 & \multicolumn{3}{c}{\cellc 46.9}   \\
      \bottomrule
    \end{tabular}
}
\label{tab:dense}
\end{table}

\subsubsection{Setup}
Beyond image classification tasks, we also validate the generalization of Conv-Adapter on dense prediction tasks, including object detection and semantic segmentation. 
We use ImageNet-21k pre-trained ResNet50 and ConvNeXt-S as backbones.
For object detection, we implement \cadapter with Faster-RCNN using the MMDetection \cite{mmdetection} framework compared with fine-tuning.
We report the average precision (AP) results on the validation split of the MS-COCO dataset \cite{lin2014microsoft}. For semantic segmentation, we implement \cadapter with UPerNet \cite{xiao2018unified} using MMSegmentation framework \cite{mmseg2020} and conduct experiments on the ADE20K dataset \cite{Zhou_2017_CVPR}, with mIoU reported on the validation split. 

For object detection, we compare all four schemes of Conv-Adapter with the fine-tuning baseline. Specifically, we follow a standard 1x training schedule: all models are trained with a batch size of 16 and optimized by AdamW with an initial learning rate of 0.0002 for Faster RCNN and 0.0001 for RetinaNet, which are then dropped by a factor of 10 at the 8-th and 11-th epoch. The shorter side of the input image is resized to 800 while maintaining the original aspect ratio.  
For segmentation, we train all models for 80k iterations with an random cropping augmentation of 512 $\times$ 512 input resolution. For ConvNeXt models, we use a larger input resolution of 640 $\times$ 640 and train the models for 160k iterations. We apply AdamW optimizer with a polynomial learning rate decay schedule. 
More detailed training setting and hyper-parameters are shown in Appendix.

\subsubsection{Results and Discussion} 
The dense prediction results are summarized in Tab. \ref{tab:dense}. We observe a different effect of Conv-Adapter on two types of backbones. On ResNet50, Conv-Adapter surpasses fine-tuning with fewer trainable parameters (including the dense prediction heads) for object detection and semantic segmentation. On ConvNeXt-S, the performance is lower than their fine-tuning counterparts. We argue that the inferior performance of Conv-Adapter on ConvNeXt-S on dense prediction tasks is due to severely reduced model capacity as the number of trainable parameters is reduced by more than 50\%. Nevertheless, they can still outperform the ResNet50 with fewer total parameters. 
This indicates there might be overfitting issues, and we encourage more future studies on this topic.

\subsection{Ablation Study} 
\label{sec:ablation}

\subsubsection{Setup} 
We provide an ablation study on the design choices of \cadapter, where we explore different architectures and adapting schemes. In this section, we mainly report the Top-1 accuracy on the validation set of VTAB-1k. 

\subsubsection{Architecture and Adapting Schemes}
We first compare the performance of Conv-Adapter using depth-wise separable, regular, and $1\times 1$ convolutions (linear layers).
As shown in Tab. \ref{tab:ablation_arch}, depth-wise separable convolution introduces the minimal parameter budget while achieving the best results. 
Apart from 4 adapting variants proposed in this work, we also explore other design choices used in previous works. We experiment on spatial down-sampling of feature maps \cite{han2020nipstinyml}. 
Compared to channel down-sampling with a bottleneck in Conv-Adapter, spatial down-sampling introduces nearly 27 times of parameters with inferior accuracy. 
We also validate the adapting scheme of applying $1 \times 1$ convolution to all convolutional layers \cite{rebufficvpr2018resadapter}, which introduces nearly 16 times of parameters to Conv-Adapter with -12.27\% accuracy gain. 
Finally, we evaluate the adapting scheme that inserts Conv-Adapter stage-wisely, which is less effective in both parameter size and performance than the proposed schemes.

\begin{table}[t!]
\centering
\caption{Ablation study on more adapting scheme and more architectures of Conv-Adapter. The different schemes and architectures mainly come from previous works. The proposed adaptation and architecture achieve the best results.}
\label{tab:ablation_arch}
\resizebox{\columnwidth}{!}{%
\begin{tabular}{l|c|c|l|l|c}
\toprule
\multicolumn{1}{c|}{Adapting Scheme} &
  \multicolumn{1}{c|}{Down-sample} &
  \# Convs &
  \multicolumn{1}{c|}{Type of Conv.} &
  \multicolumn{1}{c|}{\# Param} &
  \multicolumn{1}{c}{VTAB-1k} \\ \midrule
\cellc $K\times K$ Conv. Par. & \cellc Channel            & \cellc 2 & \cellc Depth-wise & \cellc 0.67      & \cellc \textbf{71.03}                         \\
$K\times K$ Conv. Par. & Channel            & 2 & Regular               & 5.66     & 70.52                         \\
$K\times K$ Conv. Par. & Channel            & 2 & Linear  & 1.22     & 68.32                         \\ \midrule
$K\times K$ Conv. Par. & Spatial            & 2 & Depth-wise & 18.45    & 68.54                         \\
All Conv. Par  & -                  & 1 & Linear         & 10.74  & 58.75                         \\
Stage Par.     & Channel            & 2 & Depth-wise & 1.90     & 65.06                         \\ \bottomrule
\end{tabular}%
}
\end{table}

\begin{figure}[!t]
    \centering
    \includegraphics[width=0.9\columnwidth]{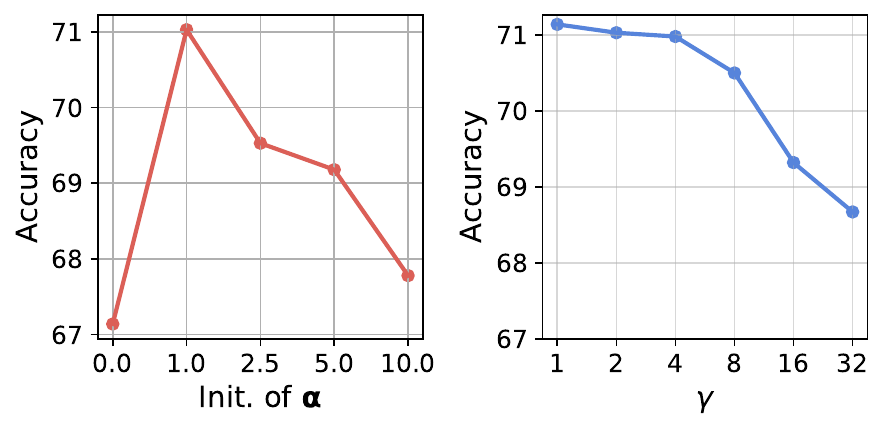}
    \caption{Sensitivity to hyper-parameters of initialization of learnable scaling vector $\boldsymbol{\alpha}$ and compression factor $\gamma$.}
    \label{fig:ablation_sensitivity}
\end{figure}

\subsubsection{Sensitivity to $\gamma$ and initialization of $\boldsymbol{\alpha}$}
We explicitly study the sensitivity of the transfer performance to the initialization of the learnable scaling vector $\boldsymbol{\alpha}$ and compression factor $\gamma$ in Conv-Adapter, as shown in Fig. \ref{fig:ablation_sensitivity}. 
When initializing $\boldsymbol{\alpha}$ as ones, Conv-Adapter achieves the best performance on the validation set of VTAB-1k. 
Compared to $\boldsymbol{\alpha}$, Conv-Adapter is more robust to the compression factor $\gamma$, achieving similar performance with the compression factor of 1, 2, and 4. Setting $\gamma$ with a larger value results in inferior performance with a more limited capacity of Conv-Adapter. 

\subsubsection{Kernel size in Conv-Adapter}

\begin{figure}
    \centering
    \includegraphics[width=0.47\columnwidth]{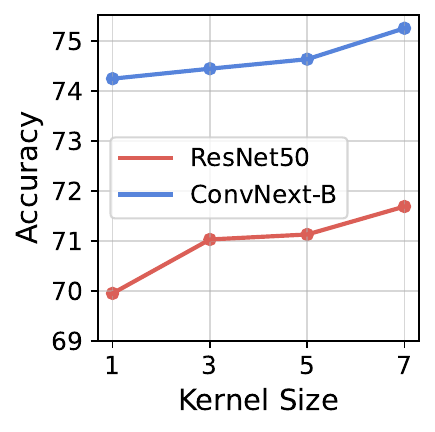}
    \caption{Sensitivity to kernel size of depth-wise convolution in Conv-Adapter, for both ResNet50 and ConvNext-B.}
    \label{fig:ablation_kernel}
\end{figure}

We show the performance of Conv-Adapter on VTAB-1k validation set in Fig \ref{fig:ablation_kernel}, of using different kernel size for the depth-wise convolution to verify our argument of the \emph{loss of locality}. One can observe that, for both ResNet50 and ConvNext-B, using smaller kernel size results in inferior performance. When setting the kernel size larger to that of the residual blocks, i.e., 5 and 7 for ResNet50, the performance is further boosted, with more parameters introduced.

\subsubsection{CKA Similarity of Conv-Adapter}
\begin{figure}[!t]
    \centering
    \includegraphics[width=0.9\columnwidth]{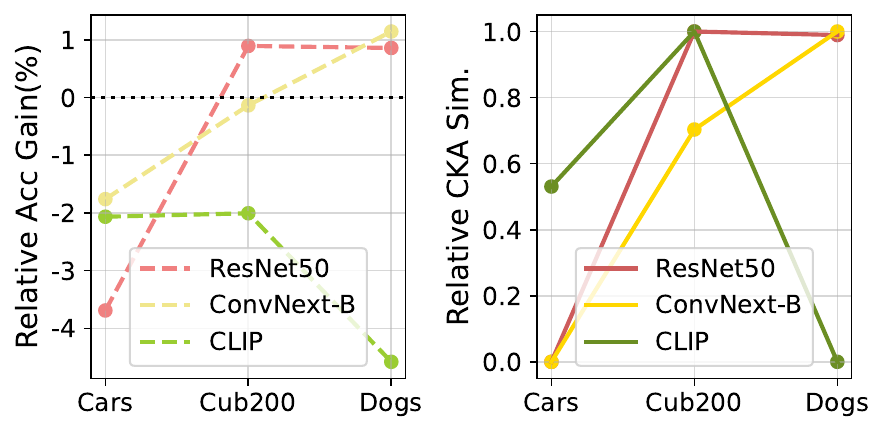}
    \caption{CKA similarity and accuracy gap between \cadapter and fully fine-tuning for FGVC datasets.}
    \label{fig:cka_acc}
\end{figure}

We observe from Tab. \ref{tab:transferability} and Tab. \ref{tab:universality} that, on datasets with large domain shifts, Conv-Adapter (and baseline methods) may fail to generalize well.
To investigate the reason, we compute the CKA similarity \cite{kornblith2019similarity,raghu2021vision} between weights of convolutional filters for the pre-trained and fine-tuned backbone. 
The lower the CKA similarity, the larger capacity is required for good transfer performance. 
We plot the CKA similarity and the relative accuracy gain of \cadapter to fine-tuning in Fig. \ref{fig:cka_acc}, where the same trends over datasets exhibit  
for different architectures. 
{When fully fine-tuning only leads to small changes in filter weights (larger CKA similarities), \cadapter is more likely to surpass the performance of fully fine-tuning.} More detail on CKA similarity comparison is in Appendix.

\section{Conclusions}
In this work, we propose \cadapter, a parameter efficient tuning module for ConvNets. \cadapter is light-weight, domain-transferable, and model-agnostic. Extensive experiments on classification and dense prediction tasks show it can achieve performance comparable to full fine-tuning with much fewer parameters. We find \cadapter might fail on tasks with large domain shifts and subject to feature quality determined by pre-training. Future work includes more exploration of \cadapter on domain robustness and dense predictions and NAS for \cadapter. 

%% file: sec/X_suppl.tex
\clearpage
\setcounter{page}{1}
\maketitlesupplementary

\section{Implementation Details}

In this section, we provide more implementation details of Conv-Adapter. We first show the details of the datasets we used and the pre-trained models we used. Then we present the details of hyper-parameter used for each method and each dataset in experiments. We implement all ConvNets and Conv-Adapter in PyTorch, and the code will be made available.

\subsection{Datasets}

The specifications of the all datasets evaluated in experiments are shown in Tab. \ref{tab:appendix_data}.

\begin{table}[!h]
\centering
\caption{Specification of all datasets evaluated. We use * to indicated randomly sampled train and validation sets (from original training set) for datasets which do not have validation split.}
\label{tab:appendix_data}
\resizebox{\columnwidth}{!}{%
\begin{tabular}{@{}l|l|l|lll@{}}
\toprule
\multicolumn{1}{c|}{Dataset}           & \multicolumn{1}{c|}{Description}  & \multicolumn{1}{c|}{\# Class}   &  \multicolumn{1}{c|}{\# Train}  & \multicolumn{1}{c|}{\# Val}  & \multicolumn{1}{c|}{\# Test}  \\ \midrule
CUB-200-2011      & \multirow{4}{*}{FGVC}     & 200      & 5,394*/5,994    & 600*    & 5,794   \\
NABirds           &                           & 700      & 21,536*/23,929   & 2,393*  & 24,633  \\
Stanford Dogs     &                           & 120      & 10,800*/12,000  & 1,200*  & 8,580   \\
Stanford Cars     &                           & 196      & 7,329*/8,144    & 815*    & 8,041   \\ \midrule
CIFAR-100      & \multirow{7}{*}{Natural}     & 100 & \multirow{7}{*}{800/200} & \multirow{7}{*}{200} & 10,000 \\
Caltech101        &                           & 102      &          &        & 6,084   \\
DTD               &                           & 47       &          &        & 1,880   \\
Oxford Flowers102 &                           & 102      &          &        & 6,149   \\
Oxford Pets       &                           & 37       &          &        & 3,669   \\
SVHN              &                           & 10       &          &        & 26,032  \\
Sun397            &                           & 397      &          &        & 21,750  \\ \midrule
Patch Camelyon & \multirow{4}{*}{Specialized} & 2   & \multirow{4}{*}{800/200} & \multirow{4}{*}{200} & 32,768 \\
EuroSAT           &                           & 10       &          &        & 5,400   \\
Resisc45          &                           & 45       &          &        & 6,300   \\
Retinopathy       &                           & 5        &          &        & 42,670  \\ \midrule
Clevr/count    & \multirow{8}{*}{Structured}  & 8   & \multirow{8}{*}{800/200} & \multirow{8}{*}{200} & 15,000 \\
Clevr/dist        &                           & 6        &          &        & 15,000  \\
DMLab             &                           & 6        &          &        & 22,735  \\
KITTI/dist        &                           & 4        &          &        & 711     \\
dSprites/loc      &                           & 16       &          &        & 73,728  \\
dSprite/ori       &                           & 16       &          &        & 73,728  \\
SmallNORB/azi     &                           & 18       &          &        & 12,150  \\
SmallNORB/ele     &                           & 9        &          &        & 12,150  \\ \midrule
FGVCAirCraft      & \multirow{5}{*}{Few-Shot} & 102      & \multirow{5}{*}{shots $\times$ classes}         &  3,333      &    3,333      \\
Food101           &                           & 101      &          &    20,200   &  30,300       \\
Oxford Flowers102 &                           & 102      &          &   1,633     & 2,463          \\
Oxford Pets       &                           & 37       &          &    736    &  3,669       \\
Stanford Cars     &                           & 196      &     & 1,635     & 8,041   \\ \midrule
MS-COCO           & Detection                 & 80      &  117,266  &  5,000  &   -    \\
ADE-20k           & Segmentation              & 150     &  25,574  &  2,000  &   -     \\ \bottomrule
\end{tabular}%
}
\end{table}

\subsection{Models}

We present the details of the pre-trained models used in experiments in Tab. \ref{tab:appendix_model}, with the checkpoint link.

\begin{table*}[!h]
\centering
\caption{Specification of pre-trained models used in experiments. }
\label{tab:appendix_model}
\resizebox{0.85 \linewidth}{!}{%
\begin{tabular}{@{}lllll@{}}
\toprule
Backbone       & Pre-trained Objective & Pre-trained Dataset & \# Param (M) & Feature Dim  \\ \midrule
ResNet50 \cite{he2016resnet}       & Supervised            & ImageNet-1k         &   23.5       &   2,048          \\
ResNet50 \cite{he2016resnet}       & Supervised            & ImageNet-21k        &   23.5       &  2,048            \\
ResNet50 BiT-M \cite{alex2019big} & Supervised            & ImageNet-21k        &     23.5     &     2,048    \\
ConvNext-B \cite{liu2022convnet}    & Supervised            & ImageNet-1k         & 87.6         & 1,024         \\
ConvNext-B \cite{liu2022convnet}    & Supervised            & ImageNet-21k        &    87.6      & 1,024         \\
ConvNext-L \cite{liu2022convnet}    & Supervised            & ImageNet-21k        &  196.2         &   1,536     \\
Swin-B  \cite{liu2021Swin}       & Supervised            & ImageNet-21k        &    86.7      &     1,024          \\
Swin-L  \cite{liu2021Swin}         & Supervised            & ImageNet-21k        &    194.9      &    1,536          \\
ResNet50 \cite{radford2021learningclip}       & CLIP                  &    CLIP                     &  38.3        &     1,024         \\
ResNet50x4 \cite{radford2021learningclip}     & CLIP                  &    CLIP                 &   87.1       &    640           \\
ResNet50  \cite{he2020moco}     & MoCov3                & ImageNet-1k       &   23.5        &    2,048     \\ \bottomrule
\end{tabular}%
}
\end{table*}

\subsection{Hyper-parameters in Experiments}

We provide the hyper-parameters search range and important settings used in experiments in this section. The detailed hyper-parameters used in experiments will be made available as configuration files in code.

\subsubsection{Classification on FGVC and VTAB-1k}

For classification tasks of FGVC and VTAB-1k, we summarize the hyper-parameter range in Tab. \ref{tab:appendix_hyper}. For VTAB-1k, we use the recommended optimal data augmentations in \cite{zhai2019largescale}, rather than solely Resize and Centre Crop as in \cite{zhang2022NOAH}. We find the recommended augmentations produces better results for full-tuning. For FGVC, we use RandomResized Crop with a minimum scale of 0.2 and Horizontal Flip \cite{2015inception} as augmentation. For few-shot classifications, we use the same range as in Tab. \ref{tab:appendix_hyper} and same augmentations as for FGVC tasks.

\begin{table}[!h]
\centering
\caption{Hyper-parameter range for grid-search on image classification tasks of FGVC and VTAB-1k.}
\label{tab:appendix_hyper}
\resizebox{0.6\columnwidth}{!}{%
\begin{tabular}{@{}ll@{}}
\toprule
             & \multicolumn{1}{c}{All Backbones}      \\ \midrule
Optimizer    & AdamW \cite{loshchilov2017decoupled} \\
LR Range     & [1e-3, 5e-4, 1e-4, 5e-5, 1e-5] \\
WD Range     & [1e-2, 1e-3, 1e-4, 0]       \\
LR schedule  & cosine \\
Total Epochs & 100    \\
Warmup       & 10     \\ \bottomrule
\end{tabular}%
}
\end{table}

\subsection{Dense Prediction Tasks}
\subsubsection{Object Detection} We compare all four schemes of Conv-Adapter with the fine-tuning baseline. Specifically, we follow a standard 1x training schedule: all models are trained with a batch size of 16 and optimized by AdamW with an initial learning rate of 0.0002 for Faster RCNN and 0.0001 for RetinaNet, which are then dropped by a factor of 10 at the 8-th and 11-th epoch. The shorter side of the input image is resized to 800 while maintaining the original aspect ratio. 

\subsubsection{Semantic Segmentation} We conduct similar experiments for the segmentation task. For ResNet50 backbones, we train all models for 80k iterations with an random cropping augmentation of 512 $\times$ 512 input resolution. For ConvNeXt models, we use a larger input resolution of 640 $\times$ 640 and train the models for 160k iterations. We apply AdamW optimizer with a polynomial learning rate decay schedule.

\section{Extended Analysis}

\subsection{Model Analysis}

In this section, we provide an analysis of the trainable parameters, model latency, and GFLOPs, based on ResNet50 \cite{he2016resnet} and ConvNext-B \cite{liu2022convnet}. Since Conv-Adapter is applied on each residual block, we first provide a theoretical analysis of the trainable parameters of each adapting scheme proposed in Tab. \ref{tab:model_analysis}.
Take the bottleneck residual block of ResNet50 as an example, we set the channel size for each convolution in the residual block as $C_{in}$, $C_{mid}$, and $C_{out}$ respectively, where $C_{in}$ is usually set to $\frac{C_{in}}{4}$. We assume the spatial size of the feature maps do not change at each residual block. 

We also provide the measurement of training/testing latency, memory cost, and GFLOPs for all the tasks evaluated in this paper, as shown in Tab. \ref{tab:complexity}. For image classification, we average the inference speed over a batch of 64. Although Conv-Adapter has increased testing latency because of the inference includes forwarding on both backbone and Conv-Adapter, the latency and memory cost of training is not necessarily greater thanks to reduced overhead of gradient computation. 

\begin{table}[!h]
\centering
\caption{Analysis of trainable parameters of the 4 proposed adapting schemes, compared to fine-tuning.}
\label{tab:model_analysis}
\resizebox{\columnwidth}{!}{%
\begin{tabular}{@{}cccc@{}}
\toprule
\multicolumn{1}{c}{Tuning} & \multicolumn{1}{c}{Input} & \multicolumn{1}{c}{Output} & \multicolumn{1}{c}{Trainable Param.}  \\ \midrule
FT  &  $C_{in} \times H \times W$ & $C_{out} \times H \times W$ & $K \times K \times C_{in} \times C_{mid} + C_{in} \times C_{mid} + C_{out} \times C_{mid}$ \\
Conv. Par  & $C_{mid} \times H \times W$ & $C_{mid} \times H \times W$ & $K \times K \times C_{mid} + \frac{C_{mid}}{\gamma} \times C_{mid} $  \\
Conv. Seq. & $C_{mid} \times H \times W$ & $C_{mid} \times H \times W$ & $K \times K \times C_{mid} + \frac{C_{mid}}{\gamma} \times C_{mid}$  \\
Res. Par.  & $C_{in} \times H \times W$ & $C_{out} \times H \times W$ & $K \times K \times C_{in} + \frac{C_{in}}{\gamma} \times C_{in}$  \\
Res. Seq.  & $C_{in} \times H \times W$ & $C_{out} \times H \times W$ & $K \times K \times C_{in} + \frac{C_{in}}{\gamma} \times C_{in}$ \\ \bottomrule
\end{tabular}%
}
\end{table}

\begin{table}[h!]
    \centering
    \caption{Evaluation of model latency, memory, and GFLOPs of 4 proposed variants for ResNet-50 and ConvNext-B on image classification, object detection, and semantic segmentation}
    \resizebox{\columnwidth}{!}{%
    \begin{tabular}{@{}c|c|rr|rr|c@{}}
    \toprule
    
      \multicolumn{7}{c}{Image Classification, Input Res. (224 $\times$ 224)  } \\ \midrule
      \multicolumn{1}{c|}{\multirow{3}{*}{Backbone}} &    
      \multicolumn{1}{c|}{\multirow{3}{*}{Tuning}} &
      \multicolumn{2}{c|}{\multirow{1}{*}{Train}} &
      \multicolumn{2}{c|}{\multirow{1}{*}{Test}} &
      \multicolumn{1}{c}{\multirow{3}{*}{GFLOPs}} \\ 
      
      \multicolumn{1}{c|}{} &
      \multicolumn{1}{c|}{} &
      \multicolumn{1}{c}{Latency} &
      \multicolumn{1}{c|}{Memory} &
      \multicolumn{1}{c}{Latency} & 
      \multicolumn{1}{c|}{Memory}  &
      \multicolumn{1}{c}{} \\
      
      \multicolumn{1}{c|}{} &
      \multicolumn{1}{c|}{} &
      \multicolumn{1}{c}{(ms/img)} &
      \multicolumn{1}{c|}{(GB)} &
      \multicolumn{1}{c}{(ms/img)} & 
      \multicolumn{1}{c|}{(GB)} &
      \multicolumn{1}{c}{} \\ \midrule
    
    \multirow{5}{*}{ResNet50-BiT}  & FT & 1.40  & 7.46 &  0.43 &  2.81  &  4.12  \\
        &   Conv. Par.  & 1.21  & 7.35 &  0.48 &  2.81  &  4.34 \\ 
        &   Conv. Seq.  & 1.24  & 7.62 &  0.54 & 2.81  &  4.34   \\
        &   Res. Par.   & 1.81  & 8.45 &  0.69 & 2.85  &   7.0 \\
        &   Res. Seq.   & 1.83  & 9.78 &  0.72 & 2.83   &  7.47 \\
     \midrule
   \multirow{5}{*}{ConvNeXt-B} & FT & 4.18  & 16.96 &  1.17 &  2.92  &  15.36  \\
        & Conv. Par.  & 4.91  & 13.52 &  1.70 &  2.98  &  17.53  \\ 
        & Conv. Seq.  & 4.94  & 14.55 & 1.70  & 2.98   &  17.53  \\
        & Res. Par.   & 4.84  & 13.50 &  1.75 & 2.98   &  17.53  \\
        & Res. Seq.   & 4.84  & 14.76 & 1.72  & 2.99   &  17.6  \\
    \midrule 
    
    \multicolumn{7}{c}{Object Detection (Test only) } \\ \midrule
      \multicolumn{1}{c|}{Backbone} &    
      \multicolumn{1}{c|}{Tuning} &
      \multicolumn{2}{c|}{Input Res.} &
      \multicolumn{2}{c|}{Latency (ms/img)} &
      \multicolumn{1}{c}{GFLOPs} \\ \midrule

    \multirow{5}{*}{ResNet50} & FT & \multicolumn{2}{c|}{\multirow{5}{*}{1280 $\times$ 800}} & \multicolumn{2}{c|}{9.38} & 84.08 \\
     &   Conv. Par.  & & & \multicolumn{2}{c|}{9.37}  & 88.61 \\ 
     &   Conv. Seq.  & & & \multicolumn{2}{c|}{11.00} & 88.61 \\
     &   Res. Par.   & & & \multicolumn{2}{c|}{17.09} & 142.89\\
     &   Res. Seq.   & & & \multicolumn{2}{c|}{16.30} & 152.54 \\ \midrule     
    \multirow{5}{*}{ConvNeXt-B} & FT & \multicolumn{2}{c|}{\multirow{5}{*}{1280 $\times$ 800}} & \multicolumn{2}{c|}{28.55} &  313.45 \\
    &   Conv. Par.  & & &  \multicolumn{2}{c|}{41.00} & 357.66  \\ 
    &   Conv. Seq.  & & &  \multicolumn{2}{c|}{41.03} & 357.66  \\
    &   Res. Par.   & & &  \multicolumn{2}{c|}{41.13} & 357.66 \\
    &   Res. Seq.   & & &  \multicolumn{2}{c|}{41.22} & 359.22 \\
    \midrule 
    
    \multicolumn{7}{c}{Semantic Segmentation (Test only) } \\ \midrule
    \multirow{5}{*}{ResNet50} & FT & \multicolumn{2}{c|}{\multirow{5}{*}{2048 $\times$ 1024}} & \multicolumn{2}{c|}{17.18} & 172.19 \\
     &   Conv. Par.  & & & \multicolumn{2}{c|}{17.21}  & 181.47  \\ 
     &   Conv. Seq.  & & & \multicolumn{2}{c|}{20.23} & 181.47  \\
     &   Res. Par.   & & & \multicolumn{2}{c|}{29.79 } &  292.63 \\
     &   Res. Seq.   & & & \multicolumn{2}{c|}{ 21.22} & 312.4 \\ \midrule     
    \multirow{5}{*}{ConvNeXt-B} & FT & \multicolumn{2}{c|}{\multirow{5}{*}{2048 $\times$ 1024}} & \multicolumn{2}{c|}{ 58.67} &  641.95 \\
    &   Conv. Par.  & & &  \multicolumn{2}{c|}{83.69} & 732.48  \\ 
    &   Conv. Seq.  & & &  \multicolumn{2}{c|}{83.77} &  732.48    \\
    &   Res. Par.   & & &  \multicolumn{2}{c|}{83.07}  & 732.48\\
    &   Res. Seq.   & & &  \multicolumn{2}{c|}{ 84.21} & 735.69 \\
    \bottomrule

    \end{tabular}}

    \label{tab:complexity}
\end{table}

\subsection{More Ablation of Conv-Adapter}




\subsection{CKA Similarity Analysis}
While the accuracy performance is well compared for PET methods, theoretical understandings towards under which circumstances PET works better than Fine-tuning lack discovery yet. 
In this section, we study how weights of backbones change with Fine-tuning using Centered Kernel Analysis and empirically discover insightful observations. 

\subsubsection{Similarity Measurement between Filter Weights using CKA}

As shown in the experimental results, whether the performance of PET surpasses Fine-tuning varies from datasets and domains. From the perspective of trainable weights, PET replaces the whole backbone with much smaller number of parameters compared with Fine-tuning. 
With the pre-trained backbone and the fine-tuned backbone, we first compute the similarity between the weights of each convolution filter using Centered Kernel Alignment (CKA). In doing so, the changes of weights brought by Fine-tuning are quantified by similarity distances between filters. 

CKA is used to compute the representation similarity between hidden layers of neural networks \cite{kornblith2019similarity,raghu2021vision}. By inputting matrices $\mathbf{X} \in \mathbb{R}^{n \times m_1}$, $\mathbf{Y} \in \mathbb{R}^{n \times m_2}$, and their Gram matrices $\mathbf{K} = \mathbf{X}\mathbf{X}^T$ and $\mathbf{L} = \mathbf{Y}\mathbf{Y}^T$, CKA follows:
\begin{align*}
    \operatorname{CKA}(\mathbf{K}, \mathbf{L}) = \frac{\operatorname{HSIC}(K,L)}{\sqrt{\operatorname{HSIC}(K,K)\operatorname{HSIC}(L,L)}},
\end{align*}
Where $\operatorname{HSIC}$ is the Hilbert-Schmidt independence criterion. 
Instead of analyzing the representation similarity, we focus on analyzing how filter weights change by Fine-tuning and utilize CKA to compute the similarity between filter weights. 
For each $k \times k$ convolution layer with $c_1$ input channels and $c_2$ output channels, weights from the initial pre-trained backbone is referred as $\mathbf{W}_i$ and weights from the fine-tuned backbone is referred as $\mathbf{W}_f$. The filter weights are reshaped into matrices for CKA computation:

\begin{itemize}
    \item For $k=1$, the convolutional filter serves as a linear transformation between channels. When computing CKA similarity, $X = \mathbf{W}_i, \mathbf{W}_i \in \mathbb{R}^{c_1 \times c_2}$ and $Y = \mathbf{W}_f, \mathbf{W}_f \in \mathbb{R}^{c_1 \times c_2}$.
    \item for $k > 1$, the weight matrix can be viewed as $c_1 \times c_2$ filters and each filter carries a size of $k \times k$ weights. When computing CKA similarity, $X = \mathbf{W}_i, \mathbf{W}_i \in \mathbb{R}^{k^2 \times c_1c_2}$, $Y = \mathbf{W}_f, \mathbf{W}_f \in \mathbb{R}^{k^2 \times c_1c_2}$.
\end{itemize}

\begin{figure}[!t]
    \centering
    \includegraphics[width=0.9\columnwidth]{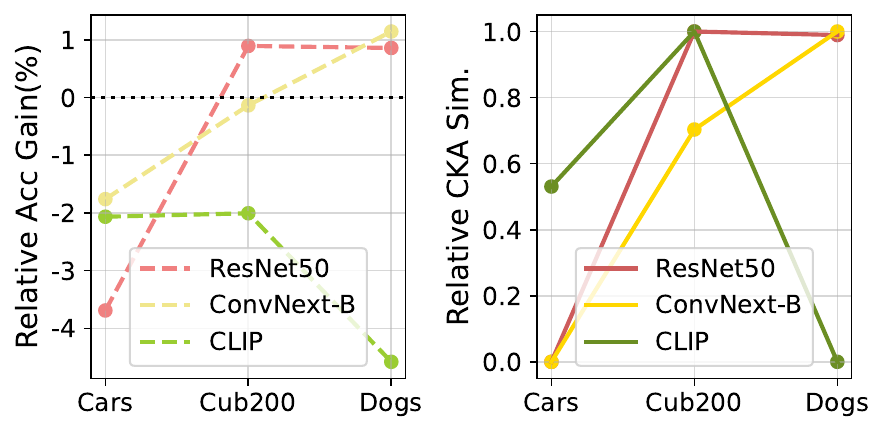}
    \caption{CKA Similarity and Accuracy gap between Conv-Adapter and fully fine-tuning for FGVC datasets. }
    \label{fig:cka_acc}
\end{figure}

For each ConvNet, we compute the average of CKA similarities among all convolutional filters and show the results of ResNet, ConvNext and ResNet-CLIP in Fig.~\ref{fig:cka_acc}. With a relatively low accuracy of Finetuning, the similarity between filter weights may not be well representative due to insufficient optimization. Thus NABirds is removed in the analysis. We also measure the domain difference between datasets with ImageNet1k using Maximum Mean Discrepancy ( Details are in the following section). Firstly we observe that
with less domain difference between target dataset and pre-trained dataset, the Conv-Adapter achieves closer performance with fully finetuning. Secondly, as shown in Fig.~\ref{fig:cka_acc}, accuracy gain of Conv-Adapter and CKA similarities between filter weights share the same trends over datasets and this phenomenon generalizes over different architectures. \textit{When fully finetuning only leads to small changes on filter weights (larger CKA similarities), Conv-Adapter is more likely to surpass the performance of fully finetuning.} 

\subsubsection{Domain Difference Quantification using MMD (Maximum Mean Discrepancy)} \label{sec:mmd}

Maximum Mean Discrepancy (MMD): measures the distance between two data distributions $p$ and $q$. 
$\phi()$ refers to a feature extractor (could be a functional intermediate layer):
\begin{equation}
    \operatorname{MMD}(p, q) = \| \mathbf{E}_{p}[\phi(\mathbf{x})] - \mathbf{E}_{q}[\phi(\mathbf{x})]\|_{\mathcal{H}_{k}}^2,
\end{equation}
where $\mathcal{H}_{k}$ refers to the kernel Hilbert space. We consider the domain difference between ImageNet1k and each dataset from FGVC. Specifically, The features subtracted from pre-trained backbones namely ResNet50 (pre-trained by Imagenet1K, ImageNet21K and CLIP), ConvNext-B (pretrained by ImageNet1K and ImageNet21K). MMD with Gaussian Kernel is computed using features from each backbone and the average MMD over all backbones is used in Fig.~\ref{fig:cka_acc}.

\section{Supplementary Results}
In this section, we provide some supplementary results to the main paper.

\subsection{Detailed Results on VTAB-1k}

We provide the per-task results on VTAB-1k on ResNet50 BiT-M \cite{alex2019big} and ConvNext-B \cite{liu2022convnet} in Tab. \ref{tab:pertask_vtab_r50} and Tab. \ref{tab:pertask_vtab_convnext} respectively. 

\begin{table*}[h!]
\centering
\caption{Per-task VTAB-1k results of ImageNet-21k pretrained ResNet50 BiT-M.}
\label{tab:pertask_vtab_r50}
\resizebox{\linewidth}{!}{%
\begin{tabular}{lrccccccccccccccccccc}
\toprule[1pt]
\multicolumn{1}{c}{Tuning} &
  \multicolumn{1}{c}{\# Param} &
  \multicolumn{1}{c}{Caltech101} &
  \multicolumn{1}{c}{Cifar100} &
  \multicolumn{1}{c}{DTD} &
  \multicolumn{1}{c}{Flowers102} &
  \multicolumn{1}{c}{Pets} &
  \multicolumn{1}{c}{Sun397} &
  \multicolumn{1}{c}{SVHN} &
  \multicolumn{1}{c}{Patch Camelyon} &
  \multicolumn{1}{c}{EuroSAT} &
  \multicolumn{1}{c}{Resisc45} &
  \multicolumn{1}{c}{Diabetic Retinopathy} &
  \multicolumn{1}{c}{Clevr/count} &
  \multicolumn{1}{c}{Clevr/dist} &
  \multicolumn{1}{c}{Dmlab} &
  \multicolumn{1}{c}{Dsprites/loc} &
  \multicolumn{1}{c}{Dsprites/ori} &
  \multicolumn{1}{c}{Kitti} &
  \multicolumn{1}{c}{Smallnorb/azi} &
  \multicolumn{1}{c}{Smallnorb/ele} \\
\midrule
FT &
  23.63 &
  84.79±0.46 &
  48.28±0.56 &
  65.32±0.3 &
  97.5±0.05 &
  86.74±0.46 &
  38.14±0.24 &
  84.57±0.91 &
  85.2±0.39 &
  95.46±0.17 &
  84.03±0.15 &
  78.74±0.11 &
  96.77±1.37 &
  58.15±0.3 &
  51.17±0.08 &
  94.39±0.96 &
  69.77±0.68 &
  78.99±0.46 &
  41.79±0.62 &
  42.74±0.19 \\
LP &
  0.11 &
  84.35±0.51 &
  44.02±0.18 &
  66.49±0.15 &
  98.85±0.03 &
  88.16±0.23 &
  43.24±0.58 &
  46.8±0.06 &
  79.88±0.33 &
  92.53±0.15 &
  78.65±0.24 &
  74.64±0.08 &
  50.43±0.09 &
  33.91±0.19 &
  37.92±0.16 &
  34.23±0.07 &
  33.67±0.09 &
  66.95±0.34 &
  18.27±0.19 &
  27.96±0.09 \\
Bias &
  0.15 &
  83.75±0.08 &
  41.99±0.4 &
  66.31±0.35 &
  97.84±0.04 &
  87.91±0.45 &
  39.29±0.21 &
  45.34±0.3 &
  79.82±0.19 &
  91.07±0.03 &
  75.77±0.62 &
  74.72±0.04 &
  41.97±0.13 &
  33.27±0.17 &
  37.86±0.03 &
  18.4±0.13 &
  19.43±0.43 &
  67.32±0.26 &
  13.59±0.23 &
  25.55±0.44 \\
VPT &
  0.15 &
  83.4±0.87 &
  34.92±0.15 &
  59.06±0.13 &
  98.1±0.38 &
  86.14±0.37 &
  43.34±0.22 &
  53.08±0.31 &
  81.06±0.99 &
  91.04±0.09 &
  75.07±0.21 &
  74.25±0.09 &
  49.2±0.43 &
  46.25±0.31 &
  38.64±0.16 &
  41.87±0.93 &
  33.53±2.25 &
  43.84±31.0 &
  20.6±0.53 &
  27.2±0.49 \\ \midrule
Conv. Par. &
  0.48 &
  85.26±0.49 &
  48.29±0.07 &
  68.79±0.44 &
  98.28±0.18 &
  86.16±0.03 &
  43.9±0.34 &
  77.55±0.18 &
  84.25±0.59 &
  95.45±0.13 &
  80.67±0.17 &
  76.48±0.21 &
  78.57±1.45 &
  49.17±0.42 &
  46.37±0.73 &
  68.3±6.06 &
  70.55±0.75 &
  78.11±0.52 &
  27.84±0.71 &
  34.69±0.22 \\
Conv. Seq. &
  0.67 &
  83.43±0.49 &
  48.92±0.38 &
  68.09±0.64 &
  97.89±0.25 &
  85.75±0.35 &
  42.78±0.17 &
  79.11±1.13 &
  84.08±0.55 &
  94.23±0.17 &
  80.78±0.53 &
  76.32±0.09 &
  73.73±2.04 &
  50.61±0.47 &
  46.16±0.2 &
  85.51±3.07 &
  71.7±0.63 &
  75.76±1.57 &
  30.52±0.24 &
  34.03±0.22 \\
Res. Par. &
  4.61 &
  85.94±0.57 &
  44.2±0.77 &
  67.29±0.67 &
  98.1±0.02 &
  86.57±0.6 &
  40.4±1.93 &
  79.75±0.61 &
  84.07±0.62 &
  94.84±0.31 &
  83.3±0.13 &
  76.59±0.09 &
  84.18±1.87 &
  54.83±1.03 &
  45.42±0.61 &
  95.78±0.23 &
  66.81±0.58 &
  76.98±0.59 &
  30.72±0.62 &
  35.97±0.43 \\
Res. Seq. &
  7.06 &
  85.4±0.49 &
  45.27±0.76 &
  65.44±0.37 &
  98.18±0.05 &
  86.21±0.17 &
  42.18±0.1 &
  79.53±0.32 &
  84.9±0.37 &
  95.38±0.12 &
  82.43±0.52 &
  76.67±0.15 &
  79.23±1.13 &
  56.54±1.45 &
  48.02±0.58 &
  96.38±0.62 &
  70.41±0.23 &
  72.85±1.21 &
  31.17±1.0 &
  36.05±0.08 \\
\bottomrule
\end{tabular}%
}
\end{table*}

\begin{table*}[!h]
\centering
\caption{Per-task VTAB-1k results of ImageNet-21k pre-trained ConvNext-B.}
\label{tab:pertask_vtab_convnext}
\resizebox{\linewidth}{!}{%
\begin{tabular}{lrccccccccccccccccccc}
\toprule[1pt]
\multicolumn{1}{c}{Tuning} &
  \multicolumn{1}{c}{\# Param} &
  Caltech101 &
  Cifar100 &
  DTD &
  Flowers102 &
  Pets &
  Sun397 &
  SVHN &
  Patch Camelyon &
  EuroSAT &
  Resisc45 &
  Diabetic Retinopathy &
  Clevr/count &
  Clevr/dist &
  Dmlab &
  Dsprites/loc &
  Dsprites/ori &
  Kitti &
  Smallnorb/azi &
  Smallnorb/ele \\
\midrule
FT &
  87.62 &
  91.97±0.69 &
  69.06±0.42 &
  76.15±0.28 &
  99.55±0.02 &
  92.12±0.26 &
  52.48±0.19 &
  89.78±0.22 &
  86.41±0.31 &
  96.08±0.16 &
  88.32±0.26 &
  78.48±0.27 &
  93.78±0.98 &
  55.9±5.55 &
  56.06±0.67 &
  96.35±0.18 &
  70.21±0.81 &
  78.44±0.74 &
  39.15±0.47 &
  36.29±0.39 \\
LP &
  0.05 &
  89.48±0.11 &
  60.53±0.28 &
  75.71±0.07 &
  99.58±0.01 &
  92.02±0.15 &
  57.44±0.17 &
  55.96±0.15 &
  83.13±0.36 &
  93.59±0.18 &
  82.78±0.3 &
  75.74±0.0 &
  55.39±0.1 &
  37.69±0.04 &
  43.1±0.07 &
  26.01±0.06 &
  37.72±0.03 &
  67.23±0.71 &
  19.94±0.1 &
  27.71±0.17 \\
Bias &
  0.18 &
  89.14±0.75 &
  61.38±0.31 &
  76.33±0.04 &
  99.65±0.02 &
  90.64±0.97 &
  51.26±0.31 &
  86.38±0.19 &
  85.76±0.32 &
  95.33±0.18 &
  83.71±0.18 &
  77.17±0.35 &
  74.3±1.65 &
  48.27±0.47 &
  52.19±0.48 &
  93.78±1.71 &
  65.5±0.83 &
  75.34±1.26 &
  31.51±0.34 &
  29.18±0.23 \\
VPT &
  0.10 &
  89.79±0.46 &
  57.8±0.23 &
  73.46±0.22 &
  99.58±0.03 &
  92.3±0.22 &
  55.55±0.1 &
  58.33±0.24 &
  83.11±0.16 &
  93.13±0.2 &
  83.01±0.12 &
  74.76±0.38 &
  58.58±0.45 &
  46.52±0.76 &
  39.0±0.47 &
  53.09±0.52 &
  27.38±3.56 &
  64.93±0.43 &
  20.75±0.46 &
  31.44±1.06 \\
\midrule
Conv. Par. &
  7.83 &
  90.94±0.32 &
  66.0±0.06 &
  74.91±0.44 &
  98.81±0.21 &
  92.4±0.18 &
  52.87±0.26 &
  88.44±0.46 &
  85.96±0.17 &
  95.61±0.08 &
  85.72±0.25 &
  77.86±0.11 &
  86.53±1.66 &
  59.48±1.19 &
  55.0±0.19 &
  93.67±0.65 &
  67.11±0.78 &
  83.5±0.88 &
  39.01±0.21 &
  34.72±0.21 \\
Conv. Seq. &
  9.58 &
  90.28±0.31 &
  68.28±0.94 &
  76.22±0.54 &
  98.48±0.09 &
  91.29±0.08 &
  53.43±0.27 &
  88.03±0.25 &
  86.32±0.05 &
  94.98±0.24 &
  85.64±0.18 &
  77.69±0.16 &
  91.17±0.7 &
  51.15±6.0 &
  52.88±0.44 &
  90.58±0.79 &
  68.22±0.24 &
  83.08±0.83 &
  38.26±0.66 &
  37.41±0.79 \\
Res. Par. &
  9.14 &
  91.41±0.9 &
  64.98±0.25 &
  73.33±0.5 &
  99.43±0.02 &
  91.66±0.28 &
  52.21±0.21 &
  88.94±0.38 &
  85.59±0.34 &
  95.51±0.13 &
  84.51±0.22 &
  77.58±0.21 &
  89.23±0.41 &
  56.34±0.99 &
  55.14±0.13 &
  90.88±0.1 &
  65.65±0.52 &
  81.25±1.19 &
  38.1±0.28 &
  37.78±0.4 \\
Res. Seq. &
  10.73 &
  89.26±1.24 &
  63.75±0.76 &
  74.61±0.33 &
  99.33±0.11 &
  90.69±0.34 &
  51.37±0.38 &
  88.47±0.45 &
  85.77±0.27 &
  95.57±0.13 &
  85.47±0.7 &
  77.72±0.12 &
  91.54±0.47 &
  52.26±1.89 &
  55.06±0.51 &
  61.9±1.12 &
  64.35±0.49 &
  82.93±0.07 &
  36.74±0.2 &
  38.72±1.47 
\\
\bottomrule
\end{tabular}%
}
\end{table*}

\subsection{Detailed Results on FGVC}

We provide the per-task results on FGVC on ResNet50 BiT-M \cite{alex2019big} and ConvNext-B \cite{liu2022convnet} in Tab. \ref{tab:pertask_fgvc_r50} and Tab. \ref{tab:pertask_fgvc_convnext} respectively. 

\begin{table}[!h]
\centering
\caption{Per-task FGVC results of ImageNet-21k pre-trained ResNet50 BiT-M.}
\label{tab:pertask_fgvc_r50}
\resizebox{0.9 \columnwidth}{!}{%
\begin{tabular}{lrcccc}
\toprule[1pt]
\multicolumn{1}{c}{Tuning} & \multicolumn{1}{c}{\# Param} & CUB200 & Stanford Dogs & Stanford Cars & NABirds \\
\midrule
FT         & 24.15 & 84.51±0.08 & 79.75±0.08 & 89.59±0.25  & 79.97±0.15 \\
LP         & 0.63  & 86.07±0.13 & 80.48±0.07 & 64.31±0.26  & 70.89±0.02 \\
Bias       & 0.67  & 79.13±0.28 & 76.49±0.11 & 34.63±0.1   & 69.68±0.11 \\
VPT        & 0.69  & 85.96±0.1  & 79.58±0.11 & 56.9±0.46   & 76.72±0.1  \\
\midrule
Conv. Par. & 1.22  & 86.41±0.2  & 82.07±0.1  & 85.78±0.25  & 80.83±0.09 \\
Conv. Seq. & 1.06  & 85.48±0.19 & 80.5±0.09  & 73.47±11.62 & 79.27±0.15 \\
Res. Par.  & 7.82  & 85.98±0.15 & 81.91±0.11 & 88.96±0.05  & 80.13±0.16 \\
Res. Seq.  & 11.80 & 85.85±0.22 & 80.69±0.01 & 87.59±0.16  & 79.68±0.12 \\
\bottomrule
\end{tabular}%
}
\end{table}

\begin{table}[!h]
\centering
\caption{Per-task FGVC results of ImageNet-21k pre-trained ConvNext-B.}
\label{tab:pertask_fgvc_convnext}
\resizebox{0.9 \columnwidth}{!}{%
\begin{tabular}{lrcccc}
\toprule[1pt]
\multicolumn{1}{c}{Tuning} & \multicolumn{1}{c}{\# Param} & CUB200 & Stanford Dogs & Stanford Cars & NABirds \\
\midrule
FT         & 87.87 & 89.31±0.18 & 87.18±0.07 & 93.43±0.24 & 88.01±0.17 \\
LP         & 0.31  & 90.46±0.02 & 89.86±0.1  & 74.96±0.06 & 85.76±0.02 \\
Bias       & 0.44  & 90.86±0.07 & 89.46±0.03 & 92.05±0.12 & 88.25±0.04 \\
VPT        & 0.37  & 89.83±0.02 & 89.95±0.12 & 74.64±0.06 & 85.69±0.05 \\
\midrule
Conv. Par. & 5.81  & 89.83±0.22 & 88.38±0.34 & 91.83±0.18 & 87.06±0.07 \\
Conv. Seq. & 3.11  & 76.5±18.24 & 86.77±0.28 & 91.32±0.23 & 87.4±0.05  \\
Res. Par.  & 5.73  & 90.09±0.08 & 88.06±0.18 & 90.78±0.14 & 86.53±0.06 \\
Res. Seq.  & 8.04  & 88.57±0.07 & 87.68±0.07 & 91.61±0.1  & 87.03±0.04 \\
\bottomrule
\end{tabular}%
}
\end{table}

%% file: main.bbl
\begin{thebibliography}{64}
\providecommand{\natexlab}[1]{#1}
\providecommand{\url}[1]{\texttt{#1}}
\expandafter\ifx\csname urlstyle\endcsname\relax
  \providecommand{\doi}[1]{doi: #1}\else
  \providecommand{\doi}{doi: \begingroup \urlstyle{rm}\Url}\fi

\bibitem[Bahng et~al.(2022)Bahng, Jahanian, Sankaranarayanan, and Isola]{bahng2022visual}
Hyojin Bahng, Ali Jahanian, Swami Sankaranarayanan, and Phillip Isola.
\newblock Exploring visual prompts for adapting large-scale models.
\newblock \emph{arXiv preprint arXiv:2203.17274}, 2022.

\bibitem[Ben~Zaken et~al.(2022)Ben~Zaken, Goldberg, and Ravfogel]{ben2022bias}
Elad Ben~Zaken, Yoav Goldberg, and Shauli Ravfogel.
\newblock Bitfit: Simple parameter-efficient fine-tuning for transformer-based masked language-models.
\newblock \emph{Proceedings of the 60th Annual Meeting of the Association for Computational Linguistics (Volume 2: Short Papers)}, 2022.

\bibitem[Bommasani et~al.(2021)Bommasani, Hudson, Adeli, Altman, Arora, et~al.]{bommasani2021opportunities}
Rishi Bommasani, Drew~A. Hudson, Ehsan Adeli, Russ Altman, Simran Arora, et~al.
\newblock On the opportunities and risks of foundation models, 2021.

\bibitem[Bossard et~al.(2014)Bossard, Guillaumin, and Van~Gool]{bossard14food101}
Lukas Bossard, Matthieu Guillaumin, and Luc Van~Gool.
\newblock Food-101 -- mining discriminative components with random forests.
\newblock In \emph{European Conference on Computer Vision}, 2014.

\bibitem[Brown et~al.(2020)Brown, Mann, Ryder, Subbiah, Kaplan, Dhariwal, Neelakantan, Shyam, Sastry, Askell, Agarwal, Herbert-Voss, Krueger, Henighan, Child, Ramesh, Ziegler, Wu, Winter, Hesse, Chen, Sigler, Litwin, Gray, Chess, Clark, Berner, McCandlish, Radford, Sutskever, and Amodei]{brown2020language}
Tom~B. Brown, Benjamin Mann, Nick Ryder, Melanie Subbiah, Jared Kaplan, Prafulla Dhariwal, Arvind Neelakantan, Pranav Shyam, Girish Sastry, Amanda Askell, Sandhini Agarwal, Ariel Herbert-Voss, Gretchen Krueger, Tom Henighan, Rewon Child, Aditya Ramesh, Daniel~M. Ziegler, Jeffrey Wu, Clemens Winter, Christopher Hesse, Mark Chen, Eric Sigler, Mateusz Litwin, Scott Gray, Benjamin Chess, Jack Clark, Christopher Berner, Sam McCandlish, Alec Radford, Ilya Sutskever, and Dario Amodei.
\newblock Language models are few-shot learners, 2020.

\bibitem[Cai et~al.(2019)Cai, Zhu, and Han]{cai2018proxylessnas}
Han Cai, Ligeng Zhu, and Song Han.
\newblock Proxyless{NAS}: Direct neural architecture search on target task and hardware.
\newblock In \emph{International Conference on Learning Representations}, 2019.

\bibitem[Cai et~al.(2020)Cai, Gan, Zhu, and Han]{han2020nipstinyml}
Han Cai, Chuang Gan, Ligeng Zhu, and Song Han.
\newblock Tinytl: Reduce memory, not parameters for efficient on-device learning.
\newblock In \emph{Advances in Neural Information Processing Systems}, pages 11285--11297. Curran Associates, Inc., 2020.

\bibitem[Chen et~al.(2019)Chen, Wang, Pang, Cao, Xiong, Li, Sun, Feng, Liu, Xu, Zhang, Cheng, Zhu, Cheng, Zhao, Li, Lu, Zhu, Wu, Dai, Wang, Shi, Ouyang, Loy, and Lin]{mmdetection}
Kai Chen, Jiaqi Wang, Jiangmiao Pang, Yuhang Cao, Yu Xiong, Xiaoxiao Li, Shuyang Sun, Wansen Feng, Ziwei Liu, Jiarui Xu, Zheng Zhang, Dazhi Cheng, Chenchen Zhu, Tianheng Cheng, Qijie Zhao, Buyu Li, Xin Lu, Rui Zhu, Yue Wu, Jifeng Dai, Jingdong Wang, Jianping Shi, Wanli Ouyang, Chen~Change Loy, and Dahua Lin.
\newblock {MMDetection}: Open mmlab detection toolbox and benchmark.
\newblock \emph{arXiv preprint arXiv:1906.07155}, 2019.

\bibitem[Chen et~al.(2017)Chen, Papandreou, Schroff, and Adam]{chen2017rethinking}
Liang-Chieh Chen, George Papandreou, Florian Schroff, and Hartwig Adam.
\newblock Rethinking atrous convolution for semantic image segmentation.
\newblock \emph{arXiv preprint arXiv:1706.05587}, 2017.

\bibitem[Chen et~al.(2022)Chen, Ge, Tong, Wang, Song, Wang, and Luo]{chen2022adaptformer}
Shoufa Chen, Chongjian Ge, Zhan Tong, Jiangliu Wang, Yibing Song, Jue Wang, and Ping Luo.
\newblock Adaptformer: Adapting vision transformers for scalable visual recognition, 2022.

\bibitem[Chen et~al.(2021)Chen, Xie, and He]{chen2021mocov3}
Xinlei Chen, Saining Xie, and Kaiming He.
\newblock An empirical study of training self-supervised vision transformers.
\newblock \emph{arXiv preprint arXiv:2104.02057}, 2021.

\bibitem[Contributors(2020)]{mmseg2020}
MMSegmentation Contributors.
\newblock {MMSegmentation}: Openmmlab semantic segmentation toolbox and benchmark.
\newblock \url{https://github.com/open-mmlab/mmsegmentation}, 2020.

\bibitem[Devlin et~al.(2018)Devlin, Chang, Lee, and Toutanova]{devlin2018bert}
Jacob Devlin, Ming-Wei Chang, Kenton Lee, and Kristina Toutanova.
\newblock Bert: Pre-training of deep bidirectional transformers for language understanding, 2018.

\bibitem[Dosovitskiy et~al.(2020)Dosovitskiy, Beyer, Kolesnikov, Weissenborn, Zhai, Unterthiner, Dehghani, Minderer, Heigold, Gelly, Uszkoreit, and Houlsby]{dosovitskiy2020image}
Alexey Dosovitskiy, Lucas Beyer, Alexander Kolesnikov, Dirk Weissenborn, Xiaohua Zhai, Thomas Unterthiner, Mostafa Dehghani, Matthias Minderer, Georg Heigold, Sylvain Gelly, Jakob Uszkoreit, and Neil Houlsby.
\newblock An image is worth 16x16 words: Transformers for image recognition at scale, 2020.

\bibitem[Elsayed et~al.(2018)Elsayed, Goodfellow, and Sohl-Dickstein]{elsayed2018adversarial}
Gamaleldin~F. Elsayed, Ian Goodfellow, and Jascha Sohl-Dickstein.
\newblock Adversarial reprogramming of neural networks, 2018.

\bibitem[Fedus et~al.(2021)Fedus, Zoph, and Shazeer]{fedus2021switch}
William Fedus, Barret Zoph, and Noam Shazeer.
\newblock Switch transformers: Scaling to trillion parameter models with simple and efficient sparsity, 2021.

\bibitem[Guo et~al.(2019)Guo, Li, Wang, and Rosing]{guo2019depthwise}
Yunhui Guo, Yandong Li, Liqiang Wang, and Tajana Rosing.
\newblock Depthwise convolution is all you need for learning multiple visual domains.
\newblock \emph{Proceedings of the AAAI Conference on Artificial Intelligence}, 33:\penalty0 8368–8375, 2019.

\bibitem[He et~al.(2022)He, Zhou, Ma, Berg-Kirkpatrick, and Neubig]{he2022towards}
Junxian He, Chunting Zhou, Xuezhe Ma, Taylor Berg-Kirkpatrick, and Graham Neubig.
\newblock Towards a unified view of parameter-efficient transfer learning.
\newblock In \emph{International Conference on Learning Representations}, 2022.

\bibitem[He et~al.(2016)He, Zhang, Ren, and Sun]{he2016resnet}
Kaiming He, Xiangyu Zhang, Shaoqing Ren, and Jian Sun.
\newblock Deep residual learning for image recognition.
\newblock \emph{2016 IEEE Conference on Computer Vision and Pattern Recognition (CVPR)}, 2016.

\bibitem[He et~al.(2020)He, Fan, Wu, Xie, and Girshick]{he2020moco}
Kaiming He, Haoqi Fan, Yuxin Wu, Saining Xie, and Ross Girshick.
\newblock Momentum contrast for unsupervised visual representation learning.
\newblock \emph{2020 IEEE/CVF Conference on Computer Vision and Pattern Recognition (CVPR)}, 2020.

\bibitem[Houlsby et~al.(2019{\natexlab{a}})Houlsby, Giurgiu, Jastrzebski, Morrone, De~Laroussilhe, Gesmundo, Attariyan, and Gelly]{houlsby2019parameter}
Neil Houlsby, Andrei Giurgiu, Stanislaw Jastrzebski, Bruna Morrone, Quentin De~Laroussilhe, Andrea Gesmundo, Mona Attariyan, and Sylvain Gelly.
\newblock Parameter-efficient transfer learning for {NLP}.
\newblock In \emph{Proceedings of the 36th International Conference on Machine Learning}, 2019{\natexlab{a}}.

\bibitem[Houlsby et~al.(2019{\natexlab{b}})Houlsby, Giurgiu, Jastrzebski, Morrone, De~Laroussilhe, Gesmundo, Attariyan, and Gelly]{pmlrv97houlsby19a}
Neil Houlsby, Andrei Giurgiu, Stanislaw Jastrzebski, Bruna Morrone, Quentin De~Laroussilhe, Andrea Gesmundo, Mona Attariyan, and Sylvain Gelly.
\newblock Parameter-efficient transfer learning for {NLP}.
\newblock In \emph{Proceedings of the 36th International Conference on Machine Learning}, pages 2790--2799. PMLR, 2019{\natexlab{b}}.

\bibitem[Howard et~al.(2017)Howard, Zhu, Chen, Kalenichenko, Wang, Weyand, Andreetto, and Adam]{howard2017mobilenets}
Andrew~G. Howard, Menglong Zhu, Bo Chen, Dmitry Kalenichenko, Weijun Wang, Tobias Weyand, Marco Andreetto, and Hartwig Adam.
\newblock Mobilenets: Efficient convolutional neural networks for mobile vision applications, 2017.

\bibitem[Hu et~al.(2021)Hu, Shen, Wallis, Allen-Zhu, Li, Wang, and Chen]{hu2021lora}
Edward Hu, Yelong Shen, Phil Wallis, Zeyuan Allen-Zhu, Yuanzhi Li, Lu Wang, and Weizhu Chen.
\newblock Lora: Low-rank adaptation of large language models, 2021.

\bibitem[Jia et~al.(2022)Jia, Tang, Chen, Cardie, Belongie, Hariharan, and Lim]{jia2022vpt}
Menglin Jia, Luming Tang, Bor-Chun Chen, Claire Cardie, Serge Belongie, Bharath Hariharan, and Ser-Nam Lim.
\newblock Visual prompt tuning.
\newblock In \emph{European Conference on Computer Vision (ECCV)}, 2022.

\bibitem[Khosla et~al.(2011)Khosla, Jayadevaprakash, Yao, and Fei-Fei]{KhoslaYaoJayadevaprakashFeiFeiFGVC2011}
Aditya Khosla, Nityananda Jayadevaprakash, Bangpeng Yao, and Li Fei-Fei.
\newblock Novel dataset for fine-grained image categorization.
\newblock In \emph{First Workshop on Fine-Grained Visual Categorization, IEEE Conference on Computer Vision and Pattern Recognition}, Colorado Springs, CO, 2011.

\bibitem[Kolesnikov et~al.(2019)Kolesnikov, Beyer, Zhai, Puigcerver, Yung, Gelly, and Houlsby]{alex2019big}
Alexander Kolesnikov, Lucas Beyer, Xiaohua Zhai, Joan Puigcerver, Jessica Yung, Sylvain Gelly, and Neil Houlsby.
\newblock Big transfer (bit): General visual representation learning, 2019.

\bibitem[Kornblith et~al.(2019)Kornblith, Norouzi, Lee, and Hinton]{kornblith2019similarity}
Simon Kornblith, Mohammad Norouzi, Honglak Lee, and Geoffrey Hinton.
\newblock Similarity of neural network representations revisited.
\newblock In \emph{International Conference on Machine Learning}, pages 3519--3529. PMLR, 2019.

\bibitem[Krause et~al.(2013)Krause, Stark, Deng, and Fei-Fei]{KrauseStarkDengFeiFei3DRR2013}
Jonathan Krause, Michael Stark, Jia Deng, and Li Fei-Fei.
\newblock 3d object representations for fine-grained categorization.
\newblock In \emph{4th International IEEE Workshop on 3D Representation and Recognition (3dRR-13)}, Sydney, Australia, 2013.

\bibitem[Lester et~al.(2021)Lester, Al-Rfou, and Constant]{lester2021powerprompt}
Brian Lester, Rami Al-Rfou, and Noah Constant.
\newblock The power of scale for parameter-efficient prompt tuning.
\newblock In \emph{Proceedings of the 2021 Conference on Empirical Methods in Natural Language Processing}, pages 3045--3059, Online and Punta Cana, Dominican Republic, 2021. Association for Computational Linguistics.

\bibitem[Li and Liang(2021)]{li2021prefixtuning}
Xiang~Lisa Li and Percy Liang.
\newblock Prefix-tuning: Optimizing continuous prompts for generation, 2021.

\bibitem[Lin et~al.(2014)Lin, Maire, Belongie, Bourdev, Girshick, Hays, Perona, Ramanan, Zitnick, and Dollár]{lin2014microsoft}
Tsung-Yi Lin, Michael Maire, Serge Belongie, Lubomir Bourdev, Ross Girshick, James Hays, Pietro Perona, Deva Ramanan, C.~Lawrence Zitnick, and Piotr Dollár.
\newblock Microsoft coco: Common objects in context, 2014.
\newblock cite arxiv:1405.0312Comment: 1) updated annotation pipeline description and figures; 2) added new section describing datasets splits; 3) updated author list.

\bibitem[Liu et~al.(2021{\natexlab{a}})Liu, Yuan, Fu, Jiang, Hayashi, and Neubig]{liu2021pretrain}
Pengfei Liu, Weizhe Yuan, Jinlan Fu, Zhengbao Jiang, Hiroaki Hayashi, and Graham Neubig.
\newblock Pre-train, prompt, and predict: A systematic survey of prompting methods in natural language processing, 2021{\natexlab{a}}.

\bibitem[Liu et~al.(2021{\natexlab{b}})Liu, Lin, Cao, Hu, Wei, Zhang, Lin, and Guo]{liu2021Swin}
Ze Liu, Yutong Lin, Yue Cao, Han Hu, Yixuan Wei, Zheng Zhang, Stephen Lin, and Baining Guo.
\newblock Swin transformer: Hierarchical vision transformer using shifted windows.
\newblock In \emph{Proceedings of the IEEE/CVF International Conference on Computer Vision (ICCV)}, 2021{\natexlab{b}}.

\bibitem[Liu et~al.(2022{\natexlab{a}})Liu, Hu, Lin, Yao, Xie, Wei, Ning, Cao, Zhang, Dong, Wei, and Guo]{liu2021swinv2}
Ze Liu, Han Hu, Yutong Lin, Zhuliang Yao, Zhenda Xie, Yixuan Wei, Jia Ning, Yue Cao, Zheng Zhang, Li Dong, Furu Wei, and Baining Guo.
\newblock Swin transformer v2: Scaling up capacity and resolution.
\newblock In \emph{International Conference on Computer Vision and Pattern Recognition (CVPR)}, 2022{\natexlab{a}}.

\bibitem[Liu et~al.(2022{\natexlab{b}})Liu, Mao, Wu, Feichtenhofer, Darrell, and Xie]{liu2022convnet}
Zhuang Liu, Hanzi Mao, Chao-Yuan Wu, Christoph Feichtenhofer, Trevor Darrell, and Saining Xie.
\newblock A convnet for the 2020s.
\newblock \emph{Proceedings of the IEEE/CVF Conference on Computer Vision and Pattern Recognition (CVPR)}, 2022{\natexlab{b}}.

\bibitem[Loshchilov and Hutter(2017)]{loshchilov2017decoupled}
Ilya Loshchilov and Frank Hutter.
\newblock Decoupled weight decay regularization, 2017.

\bibitem[Maji et~al.(2013)Maji, Kannala, Rahtu, Blaschko, and Vedaldi]{maji13finegrained}
S. Maji, J. Kannala, E. Rahtu, M. Blaschko, and A. Vedaldi.
\newblock Fine-grained visual classification of aircraft.
\newblock Technical report, 2013.

\bibitem[Mallya et~al.(2018)Mallya, Davis, and Lazebnik]{mallya2018piggyback}
Arun Mallya, Dillon Davis, and Svetlana Lazebnik.
\newblock Piggyback: Adapting a single network to multiple tasks by learning to mask weights, 2018.

\bibitem[Mudrakarta et~al.(2018)Mudrakarta, Sandler, Zhmoginov, and Howard]{mudrakarta2018k}
Pramod~Kaushik Mudrakarta, Mark Sandler, Andrey Zhmoginov, and Andrew Howard.
\newblock K for the price of 1: Parameter-efficient multi-task and transfer learning, 2018.

\bibitem[Nilsback and Zisserman(2006)]{nilsback2006visual}
M-E Nilsback and Andrew Zisserman.
\newblock A visual vocabulary for flower classification.
\newblock In \emph{2006 IEEE Computer Society Conference on Computer Vision and Pattern Recognition (CVPR'06)}, pages 1447--1454. IEEE, 2006.

\bibitem[Parkhi et~al.(2012)Parkhi, Vedaldi, Zisserman, and Jawahar]{parkhi12a}
Omkar~M. Parkhi, Andrea Vedaldi, Andrew Zisserman, and C.~V. Jawahar.
\newblock Cats and dogs.
\newblock In \emph{IEEE Conference on Computer Vision and Pattern Recognition}, 2012.

\bibitem[Pfeiffer et~al.(2020)Pfeiffer, R{\"u}ckl{\'e}, Poth, Kamath, Vuli{\'c}, Ruder, Cho, and Gurevych]{pfeiffer2020adapterhub}
Jonas Pfeiffer, Andreas R{\"u}ckl{\'e}, Clifton Poth, Aishwarya Kamath, Ivan Vuli{\'c}, Sebastian Ruder, Kyunghyun Cho, and Iryna Gurevych.
\newblock {A}dapter{H}ub: A framework for adapting transformers.
\newblock In \emph{Proceedings of the 2020 Conference on Empirical Methods in Natural Language Processing: System Demonstrations}, pages 46--54, Online, 2020. Association for Computational Linguistics.

\bibitem[Pfeiffer et~al.(2021)Pfeiffer, Kamath, Rücklé, Cho, and Gurevych]{pfeiffer2021adapterfusion}
Jonas Pfeiffer, Aishwarya Kamath, Andreas Rücklé, Kyunghyun Cho, and Iryna Gurevych.
\newblock Adapterfusion: Non-destructive task composition for transfer learning.
\newblock \emph{Proceedings of the 16th Conference of the European Chapter of the Association for Computational Linguistics: Main Volume}, 2021.

\bibitem[Radford et~al.(2021)Radford, Kim, Hallacy, Ramesh, Goh, Agarwal, Sastry, Askell, Mishkin, Clark, Krueger, and Sutskever]{radford2021learningclip}
Alec Radford, Jong~Wook Kim, Chris Hallacy, Aditya Ramesh, Gabriel Goh, Sandhini Agarwal, Girish Sastry, Amanda Askell, Pamela Mishkin, Jack Clark, Gretchen Krueger, and Ilya Sutskever.
\newblock Learning transferable visual models from natural language supervision, 2021.

\bibitem[Radosavovic et~al.(2020)Radosavovic, Kosaraju, Girshick, He, and Dollar]{ilija2020regnet}
Ilija Radosavovic, Raj~Prateek Kosaraju, Ross Girshick, Kaiming He, and Piotr Dollar.
\newblock Designing network design spaces.
\newblock \emph{2020 IEEE/CVF Conference on Computer Vision and Pattern Recognition (CVPR)}, 2020.

\bibitem[Raghu et~al.(2021)Raghu, Unterthiner, Kornblith, Zhang, and Dosovitskiy]{raghu2021vision}
Maithra Raghu, Thomas Unterthiner, Simon Kornblith, Chiyuan Zhang, and Alexey Dosovitskiy.
\newblock Do vision transformers see like convolutional neural networks?
\newblock \emph{Advances in Neural Information Processing Systems}, 34:\penalty0 12116--12128, 2021.

\bibitem[Rebuffi et~al.(2018)Rebuffi, Bilen, and Vedaldi]{rebufficvpr2018resadapter}
Sylvestre-Alvise Rebuffi, Hakan Bilen, and Andrea Vedaldi.
\newblock Efficient parametrization of multi-domain deep neural networks.
\newblock In \emph{CVPR}, 2018.

\bibitem[Shelhamer et~al.(2017)Shelhamer, Long, and Darrell]{shelhamer2017fully}
Evan Shelhamer, Jonathan Long, and Trevor Darrell.
\newblock Fully convolutional networks for semantic segmentation.
\newblock \emph{IEEE transactions on pattern analysis and machine intelligence}, 39\penalty0 (4):\penalty0 640--651, 2017.

\bibitem[Szegedy et~al.(2015{\natexlab{a}})Szegedy, Liu, Jia, Sermanet, Reed, Anguelov, Erhan, Vanhoucke, and Rabinovich]{2015inception}
Christian Szegedy, Wei Liu, Yangqing Jia, Pierre Sermanet, Scott Reed, Dragomir Anguelov, Dumitru Erhan, Vincent Vanhoucke, and Andrew Rabinovich.
\newblock Going deeper with convolutions.
\newblock \emph{2015 IEEE Conference on Computer Vision and Pattern Recognition (CVPR)}, 2015{\natexlab{a}}.

\bibitem[Szegedy et~al.(2015{\natexlab{b}})Szegedy, Liu, Jia, Sermanet, Reed, Anguelov, Erhan, Vanhoucke, and Rabinovich]{szegedy2015going}
Christian Szegedy, Wei Liu, Yangqing Jia, Pierre Sermanet, Scott Reed, Dragomir Anguelov, Dumitru Erhan, Vincent Vanhoucke, and Andrew Rabinovich.
\newblock Going deeper with convolutions.
\newblock In \emph{Proceedings of the IEEE conference on computer vision and pattern recognition}, pages 1--9, 2015{\natexlab{b}}.

\bibitem[Tan and Le(2019)]{tan2019efficientnet}
Mingxing Tan and Quoc~V. Le.
\newblock Efficientnet: Rethinking model scaling for convolutional neural networks, 2019.

\bibitem[Tan and Le(2021)]{tan2021efficientnetv2}
Mingxing Tan and Quoc~V. Le.
\newblock Efficientnetv2: Smaller models and faster training, 2021.

\bibitem[Thrun(1998)]{thrun1998lifelong}
Sebastian Thrun.
\newblock Lifelong learning algorithms.
\newblock In \emph{Learning to learn}, pages 181--209. Springer, 1998.

\bibitem[Van~Horn et~al.(2015)Van~Horn, Branson, Farrell, Haber, Barry, Ipeirotis, Perona, and Belongie]{van2015building}
Grant Van~Horn, Steve Branson, Ryan Farrell, Scott Haber, Jessie Barry, Panos Ipeirotis, Pietro Perona, and Serge Belongie.
\newblock Building a bird recognition app and large scale dataset with citizen scientists: The fine print in fine-grained dataset collection.
\newblock In \emph{Proceedings of the IEEE Conference on Computer Vision and Pattern Recognition}, pages 595--604, 2015.

\bibitem[Vaswani et~al.(2017)Vaswani, Shazeer, Parmar, Uszkoreit, Jones, Gomez, Kaiser, and Polosukhin]{vaswani2017attention}
Ashish Vaswani, Noam Shazeer, Niki Parmar, Jakob Uszkoreit, Llion Jones, Aidan~N. Gomez, Lukasz Kaiser, and Illia Polosukhin.
\newblock Attention is all you need, 2017.

\bibitem[Wah et~al.(2011)Wah, Branson, Welinder, Perona, and Belongie]{WahCUB2002011}
C. Wah, S. Branson, P. Welinder, P. Perona, and S. Belongie.
\newblock The caltech-ucsd birds200-2011 dataset.
\newblock Technical Report CNS-TR-2011-001, California Institute of Technology, 2011.

\bibitem[Xiao et~al.(2018)Xiao, Liu, Zhou, Jiang, and Sun]{xiao2018unified}
Tete Xiao, Yingcheng Liu, Bolei Zhou, Yuning Jiang, and Jian Sun.
\newblock Unified perceptual parsing for scene understanding.
\newblock In \emph{Proceedings of the European Conference on Computer Vision (ECCV)}, pages 418--434, 2018.

\bibitem[Yang et~al.(2022)Yang, Rakin, and Fan]{yang2022repnet}
Li Yang, Adnan~Siraj Rakin, and Deliang Fan.
\newblock Rep-net: Efficient on-device learning via feature reprogramming.
\newblock In \emph{Proceedings of the IEEE/CVF Conference on Computer Vision and Pattern Recognition (CVPR)}, pages 12277--12286, 2022.

\bibitem[Zhai et~al.(2019)Zhai, Puigcerver, Kolesnikov, Ruyssen, Riquelme, Lucic, Djolonga, Pinto, Neumann, Dosovitskiy, Beyer, Bachem, Tschannen, Michalski, Bousquet, Gelly, and Houlsby]{zhai2019largescale}
Xiaohua Zhai, Joan Puigcerver, Alexander Kolesnikov, Pierre Ruyssen, Carlos Riquelme, Mario Lucic, Josip Djolonga, Andre~Susano Pinto, Maxim Neumann, Alexey Dosovitskiy, Lucas Beyer, Olivier Bachem, Michael Tschannen, Marcin Michalski, Olivier Bousquet, Sylvain Gelly, and Neil Houlsby.
\newblock A large-scale study of representation learning with the visual task adaptation benchmark, 2019.

\bibitem[Zhang et~al.(2020)Zhang, Wu, Zhang, Zhu, Lin, Zhang, Sun, He, Mueller, Manmatha, Li, and Smola]{zhang2020resnest}
Hang Zhang, Chongruo Wu, Zhongyue Zhang, Yi Zhu, Haibin Lin, Zhi Zhang, Yue Sun, Tong He, Jonas Mueller, R. Manmatha, Mu Li, and Alexander Smola.
\newblock Resnest: Split-attention networks, 2020.

\bibitem[Zhang et~al.(2021)Zhang, Li, Chen, Deng, Bi, Tan, Huang, and Chen]{zhang2021differentiable}
Ningyu Zhang, Luoqiu Li, Xiang Chen, Shumin Deng, Zhen Bi, Chuanqi Tan, Fei Huang, and Huajun Chen.
\newblock Differentiable prompt makes pre-trained language models better few-shot learners, 2021.

\bibitem[Zhang et~al.(2022)Zhang, Zhou, and Liu]{zhang2022NOAH}
Yuanhan Zhang, Kaiyang Zhou, and Ziwei Liu.
\newblock Neural prompt search, 2022.

\bibitem[Zhou et~al.(2017)Zhou, Zhao, Puig, Fidler, Barriuso, and Torralba]{Zhou_2017_CVPR}
Bolei Zhou, Hang Zhao, Xavier Puig, Sanja Fidler, Adela Barriuso, and Antonio Torralba.
\newblock Scene parsing through ade20k dataset.
\newblock In \emph{Proceedings of the IEEE Conference on Computer Vision and Pattern Recognition (CVPR)}, 2017.

\end{thebibliography}
